\documentclass[10pt,sigconf]{acmart}
\pdfoutput=1

\usepackage{booktabs} % For formal tables
\usepackage{multirow, makecell}
\usepackage{amsmath}
\usepackage{graphicx,epsfig,color}
\usepackage{balance}
\usepackage{comment}
\usepackage{soul}
\usepackage{color}
\usepackage{amsfonts}
\usepackage{url}
\usepackage{subfigure}
\usepackage{tablefootnote}
\usepackage[font={small},labelfont={bf,up}]{caption}
\usepackage[T1]{fontenc}
\usepackage{textcomp}
\usepackage{gensymb}
\usepackage{colortbl}

\setcopyright{acmcopyright}
\begin{document}
%\title{SIG Proceedings Paper in LaTeX Format}
%\titlenote{Produces the permission block, and
%  copyright information}
%\subtitle{Extended Abstract}
%\subtitlenote{The full version of the author's guide is available as
%  \texttt{acmart.pdf} document}

%\title{{\LARGE {\textsf {\textbf{NestDNN}}}}: Enabling Resource-Aware Concurrent \\ On-Device Deep Learning for Mobile Vision Systems}

\copyrightyear{2018} 
\acmYear{2018} 
\setcopyright{acmcopyright}
\acmConference[MobiCom '18]{The 24th Annual International Conference on Mobile Computing and Networking}{October 29-November 2, 2018}{New Delhi, India}
\acmBooktitle{The 24th Annual International Conference on Mobile Computing and Networking (MobiCom '18), October 29-November 2, 2018, New Delhi, India}
\acmPrice{15.00}
\acmDOI{10.1145/3241539.3241559}
\acmISBN{978-1-4503-5903-0/18/10}

\fancyhead{}

\title{NestDNN: Resource-Aware Multi-Tenant On-Device Deep Learning for Continuous Mobile Vision}

%\author{Biyi Fang}
%\authornote{Authors contributed equally.}
%\affiliation{
% \institution{Michigan State University}
%}
%
% \author{Xiao Zeng} 
%\affiliation{
% \institution{Michigan State University}
%}
%
% \author{Mi Zhang} 
%\affiliation{
% \institution{Michigan State University}
% \vspace{10mm}
%}

%
%\author{Biyi Fang\textsuperscript{*}, Xiao Zeng\textsuperscript{*}, Mi Zhang}
%\authornote{Authors contributed equally.}
%\affiliation{
%\institution{Michigan State University}
% \vspace{8mm}
%}

%\institution{\textsuperscript{*}Co-primary authors}
%\vspace{10mm}

\author{Biyi Fang\textsuperscript{$\dagger$}, Xiao Zeng\textsuperscript{$\dagger$}, Mi Zhang}
\affiliation{
\vspace{1.5mm}
\institution{Michigan State University}
 \vspace{0.5mm}
\institution{\textsuperscript{$\dagger$}Co-primary authors}
\vspace{3mm}
}

\newcommand{\biyi}[1]{\sethlcolor{yellow}\hl{[Biyi: #1]}}
\newcommand{\xiao}[1]{\sethlcolor{cyan}\hl{[Xiao: #1]}}
\newcommand{\mi}[1]{\sethlcolor{red}\hl{[Mi: #1]}}

\newcommand{\sysname}{\textsf{NestDNN}}
\newcommand{\systitle}{NestDNN}
\newcommand{\vgg}{VGGNet}
\newcommand{\resnet}{ResNet}

\newcommand{\squishlist}{
	\begin{list}{$\bullet$}
		{ \setlength{\itemsep}{0pt}      \setlength{\parsep}{3pt}
			\setlength{\topsep}{3pt}       \setlength{\partopsep}{0pt}
			\setlength{\leftmargin}{1.5em} \setlength{\labelwidth}{1em}
			\setlength{\labelsep}{0.5em} } }
	
	\newcommand{\squishend}{
\end{list}  }

\renewcommand{\shortauthors}{B. Fang et al.}

\begin{abstract}
	% max of 250 words
%!TEX root = mobicom2018.tex
%Recently, tremendous efforts have been made on enabling on-device deep learning to run deep learning models locally on mobile system.
%
%Although equipped with powerful CPU/GPU, t
%
Mobile vision systems such as smartphones, drones, and augmented-reality headsets are revolutionizing our lives.
These systems usually run multiple applications concurrently and their available resources at runtime are dynamic due to events such as starting new applications, closing existing applications, and application priority changes.
In this paper, we present {\sysname}, a framework that takes the \textit{dynamics of runtime resources} into account to enable \textit{resource-aware} multi-tenant on-device deep learning for mobile vision systems.
{\sysname} enables each deep learning model to offer \textit{flexible} resource-accuracy trade-offs.
At runtime, it \textit{dynamically} selects the \textit{optimal} resource-accuracy trade-off for each deep learning model to fit the model's resource demand to the system's available runtime resources.
In doing so, {\sysname} efficiently utilizes the limited resources in mobile vision systems to jointly maximize the performance of all the concurrently running applications.
Our experiments show that compared to the resource-agnostic status quo approach, {\sysname} achieves as much as $4.2\%$ increase in inference accuracy, $2.0 \times$ increase in video frame processing rate and $1.7\times$ reduction on energy consumption.
\end{abstract}





\maketitle

% 3 pages
%\vspace{-2mm}
%&latex 
%!TEX root = mobicom2018.tex

\section{Introduction}
\label{sec.intro}

%% P1: Talk about mobile vision systems are revolutionary and are the future.
%
Mobile systems with onboard video cameras such as smartphones, drones, wearable cameras, and augmented-reality headsets are revolutionizing the way we live, work, and interact with the world.
%Mobile systems with onboard video cameras such as smartphones, drones, wearable cameras, and head-mounted augmented reality devices are pervasive today.
%
By processing the streaming video inputs, these mobile systems are able to retrieve visual information from the world and are promised to open up a wide range of new applications and services.
For example, a drone that can detect vehicles, identify road signs, and track traffic flows will enable mobile traffic surveillance with aerial views that traditional traffic surveillance cameras positioned at fixed locations cannot provide \cite{puri2005survey}. 
A wearable camera that can recognize everyday objects, identify people, and understand the surrounding environments can be a life-changer for the blind and visually impaired individuals \cite{nvidia-blog}.
%For all these application scenarios, the commonality is the need to supporting concurrently running multiple computer vision algorithms with each of which specialized on different vision tasks. execute different vision tasks on streaming video frames.
%By leveraging these breakthroughs, mobile systems equipped with video cameras are capable of continuously analyzing streaming images to retrieve information from the physical world, turning the envisioned continuous mobile vision into reality \cite{bahl2012vision, likamwa2015starfish, Naderiparizi2017}.

%%%%%%%%%%%%%%%%%%%%%%%%%%%%%%%%%%%%%%%%%%%%%%%%%%%%%%%%%%%%%%%%%

%% P2: Talk about the unique challenges of mobile vision systems. 
% 1) => we need to run concurrent CV algorithms due to the complexity of video frames. % flexible (multi-task learning is not flexible)
% 2) => the load new models/page out old models due to context change. 
%
The key to achieving the full promise of these mobile vision systems is effectively analyzing the streaming video frames.
%
%However, processing streaming video frames taken in the mobile setting can be very challenging.
However, processing streaming video frames taken in mobile settings is challenging in two folds.
First, the processing usually involves \textit{multiple} computer vision tasks.
This multi-tenant characteristic requires mobile vision systems to \textit{concurrently} run multiple applications that target different vision tasks \cite{likamwa2015starfish}.
%This requires mobile vision systems to concurrently run multiple computer vision algorithms that target different vision tasks \cite{likamwa2015starfish}.
%
Second, the \textit{context} in mobile settings can be frequently changed.
%This requires mobile vision systems to be able to shift computer vision algorithms to execute new vision tasks encountered in the new context \cite{mcdnn}.
%
This requires mobile vision systems to be able to switch applications to execute new vision tasks encountered in the new context \cite{mcdnn}.
%Second, due to various accuracy, memory footprint and latency demand required by vision tasks, vision algorithms should be able to provide different model variants \cite{mcdnn} to meet the requirements and switch between them at the cost of least overhead.
%Finally, mobile runtime resources, together with mobile context, are changed very frequently.
%Mobile vision systems should promptly response to these variations, adapt its running process and allocate computing resources optimally to deliver maximum performance.

%%%%%%%%%%%%%%%%%%%%%%%%%%%%%%%%%%%%%%%%%%%%%%%%%%%%%%%%%%%%%%%%%

%% P3: Talk about 
% 1) the superiority of DNNs on vision tasks.
% 2) introduce on-device deep learning.
% 3) why on-device DL works the best for mobile vision systems.
%
In the past few years, deep learning (e.g., Deep Neural Networks (DNNs)) \cite{lecun2015deep} has become the dominant approach in computer vision due to its capability of achieving impressively high accuracies on a variety of important vision tasks  \cite{krizhevsky2012imagenet,taigman2014deepface,zhou2014learning}.
%such as object recognition \cite{krizhevsky2012imagenet}, face detection \cite{taigman2014deepface}, and scene understanding \cite{zhou2014learning}. 
%Considering the significant potential of mobile vision systems,
%
As deep learning chipsets emerge, there is a significant interest in leveraging the on-device computing resources to execute deep learning models on mobile systems without cloud support \cite{google-clips,apple-facedetection,amazon-deeplens}.
% which has access to virtually unlimited amount of resources
Compared to the cloud, mobile systems are constrained by limited resources. 
Unfortunately, deep learning models are known to be resource-demanding \cite{sze2017efficient}.
To enable on-device deep learning, one of the common techniques used by application developers is compressing the deep learning model to reduce its resource demand at a modest loss in accuracy as trade-off \cite{mcdnn, zhang2017live}.
Although this technique has gained considerable popularity and has been applied to developing state-of-the-art mobile deep learning systems \cite{howard2017mobilenets, deepmon, zeng2017mobiledeeppill, lane2016deepx, fang2017deepasl}, it has a \textit{key drawback}: 
since application developers develop their applications independently, the resource-accuracy trade-off of the compressed model is \textit{predetermined} based on a \textit{static} resource budget at application development stage and is \textit{fixed} after the application is deployed. 
However, the available resources in mobile vision systems at runtime are always \textit{dynamic} due to events such as starting new applications, closing existing applications, and application priority changes.
% the concurrently running applications change depending on the context.
%As a consequence,
As such, when the resources available at runtime do not meet the resource demands of the compressed models, resource contention among concurrently running applications occurs, forcing the streaming video to be processed at a much lower frame rate. 
On the other hand, when extra resources at runtime become available, the compressed models cannot utilize the extra available resources to regain their sacrificed accuracies back.
In this work, we present {\sysname}, a framework that takes the \textit{dynamics of runtime resources} into consideration to enable \textit{resource-aware} multi-tenant on-device deep learning for mobile vision systems.
%In this work, we propose a new paradigm / resource-aware framework to make the dynamic accuracy-resource trade-off into reality
%Although the seed model obtained via DNN pruning has an optimized model in terms of efficiency, it suffers a graceful drop of accuracy and only offers a fixed trade-off between model performance and inference speed, and hence does not provide enough flexibility to meet various requirements in a mobile environment. 
%{\sysname} not only supports concurrent deep learning models, but also allows them to share resources.
%{\sysname} takes the dynamics of runtime resources in mobile vision systems into account. 
%
{\sysname} replaces fixed resource-accuracy trade-offs with \textit{flexible} resource-accuracy trade-offs, and \textit{dynamically} selects the \textit{optimal} resource-accuracy trade-off for each deep learning model at runtime to fit the model's resource demand to the system's available runtime resources.
In doing so, {\sysname} is able to efficiently utilize the limited resources in the mobile system to jointly maximize the performance of all the concurrently running applications.

%%%%%%%%%%%%%%%%%%%%%%%%%%%%%%%%%%%%%%%%%%%%%%%%%%%%%%%%%%%%%%%%%

\vspace{1mm}
\noindent
\textbf{Challenges and our Solutions}.
%%major technical challenges for designing
The design of {\sysname} involves two key technical challenges.
(i) The limitation of existing approaches is rooted in the constraint where the trade-off between resource demand and accuracy of a compressed deep learning model is fixed.
Therefore, the first challenge lies in designing a scheme that enables a deep learning model to provide flexible resource-accuracy trade-offs.
One naive approach is to have all the possible model variants with various resource-accuracy trade-offs installed in the mobile system. 
However, since these model variants are \textit{independent} of each other, this approach is \textit{not scalable} and becomes infeasible when the mobile system concurrently runs multiple deep learning models, with each of which having multiple model variants.
(ii) Selecting which resource-accuracy trade-off for each of the concurrently running deep learning models is \textit{not trivial}.
% how to select the allocate runtime resources to different 
%
This is because different applications have different goals on inference accuracies and processing latencies.
%(i.e., approximate car counts)
Taking a traffic surveillance drone as an example: an application that counts vehicles to detect traffic jams does not need high accuracy but requires low latency; an application that reads license plates needs high plate reading accuracy but does not require real-time response \cite{zhang2017live}.

To address the first challenge, {\sysname} employs a novel \textit{model pruning and recovery} scheme which transforms a deep learning model into a single compact \textit{multi-capacity model}. 
%with multiple capacities.
%The multi-capacity model has two key features.
%First, the multi-capacity model is comprised of a set of sub-models. 
The multi-capacity model is comprised of a set of descendent models, each of which offers a unique resource-accuracy trade-off.
%Each descendent model has a unique resource-accuracy trade-off.
%
Unlike traditional model variants that are independent of each other, the descendent model with smaller capacity (i.e., resource demand) \textit{shares} its model parameters with the descendent model with larger capacity, making itself \textit{nested} inside the descendent model with larger capacity without taking extra memory space.
%the sub-model with smaller capacity is \textit{encapsulated} in the sub-model with larger capacity. 
%
In doing so, the multi-capacity model is able to provide various resource-accuracy trade-offs with a compact memory footprint.
To address the second challenge, {\sysname} encodes the inference accuracy and processing latency of each descendent model of each concurrently running application into a \textit{cost function}.
Given all the cost functions, {\sysname} employs a \textit{resource-aware runtime scheduler} which selects the optimal resource-accuracy trade-off for each deep learning model and determines the optimal amount of runtime resources to allocate to each model to jointly maximize the overall inference accuracy and minimize the overall processing latency of all the concurrently running applications.

\begin{figure*}[t]
\centering
\includegraphics[scale=0.50]{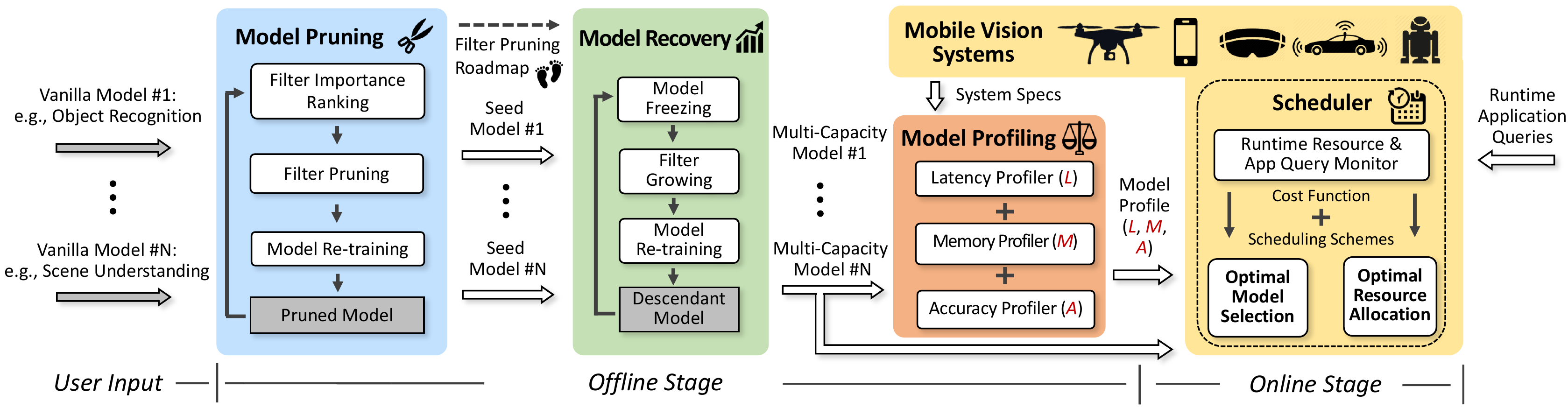}
\vspace{-1mm}
\caption{{\sysname} architecture.}
\vspace{-2mm}
\label{dia.sys1}
\end{figure*} 

%%%%%%%%%%%%%%%%%%%%%%%%%%%%%%%%%%%%%%%%%%%%%%%%%%%%%%%%%%%%%%%%%

%% P7: Summarize the key contributions of our work.
\vspace{1mm}
\noindent
\textbf{Summary of Experimental Results}.
We have conducted a rich set of experiments to evaluate the performance of {\sysname}.
%among the most important tasks for mobile vision systems
To evaluate the performance of the multi-capacity model, we evaluated it on six mobile vision applications that target some of the most important tasks for mobile vision systems.
These applications are developed based on two widely used deep learning models -- VGG Net~\cite{simonyan2014very} and ResNet~\cite{he2016deep} -- and six datasets that are commonly used in computer vision community.
To evaluate the performance of the resource-aware runtime scheduler, we incorporated two widely used scheduling schemes and implemented {\sysname} and the six mobile vision applications on three smartphones. 
We also implemented the status quo approach which uses \textit{fixed} resource-accuracy trade-off and is thus \textit{resource-agnostic}.
To compare the performance between our resource-aware approach with the resource-agnostic status quo approach, we have designed a benchmark that emulates runtime application queries in diverse scenarios.
Our results show that:
 
%
%, and ran a benchmark that contains XXX.
%The detailed contributions and important results are summarized as follows:
%\vspace{1.0mm}
%\noindent
%\textbf{Contributions.}
%
%In sum, this paper makes the following key contributions:
%
\vspace{-0mm}
\squishlist{
\item{
%\textbf{Multi-Capacity Model.}
%We have designed a pruning and recovery scheme to transform an off-the-shelf deep learning model into a single compact multi-capacity model that is able to provide optimized resource-accuracy trade-offs.
%
%The pruning and recovery scheme consists of a \textit{model pruning phase} and a \textit{model recovery phase}.
%
%For model pruning, we have devised a state-of-the-art \textit{filter pruning} approach named \textit{Triplet Response Residual} (TRR) that significant reduces not only the size of a deep learning model but also its computational cost.
%
%For model recovery, we have devised an innovative \textit{model freezing and filter growing} (i.e., \textit{freeze-\&-grow}) approach that generates the multi-capacity model in an iterative manner.
%
The multi-capacity model is able to provide flexible and optimized resource-accuracy trade-offs nested in a single model. 
%and significantly reduce model memory footprint as well as model switching overhead.
%
With parameter sharing, it significantly reduces model memory footprint and model switching overhead.
} 
%
%\item{
%\textbf{Freeze-\&-Grow for DNN Recovery.}
%We propose a novel freeze-and-grow scheme for DNN recovery.
%We have shown that we can save XXX memory, and generate multi-capability DNN models.
%
%}
%
\item{
%\textbf{Resource-Aware Scheduler.}
%We have designed a runtime scheduler that jointly maximizes the accuracy and minimizes the inference latency of concurrent vision applications running on mobile vision systems. 
%
The resource-aware runtime scheduler outperforms the resource-agnostic counterpart on both scheduling schemes, achieving as much as $4.2\%$ increase in inference accuracy, $2.0 \times$ increase in video frame processing rate and $1.7\times$ reduction on energy consumption.
%We have shown that {\sysname} can achieve up to $3.2 \times$ FPS and $13.5\%$ increase in accuracy.
%Our results show that MCDNN can make effective trade-offs between resource utilization and accuracy (e.g., transform models to use 4 fewer FLOPs and roughly 5less memory at 1-4% loss in accuracy), share models across DNNs with significant savings (e.g., 1-4 orders of magnitude less memory use and 1.2-100less compute/energy), effectively specialize models to various contexts (e.g., specialize models with 5-25 less compute, two orders of magnitude less storage, and still achieve accuracy gains), and schedule computations across both mobile and cloud devices for diverse operating conditions (e.g., disconnected operation, low resource availability, and varying number of applications).
}
%\item{
%\textbf{Enabling Resource-Aware Concurrent On-Device Deep Learning for Mobile Vision Systems.}
%}
}\squishend

%%%%%%%%%%%%%%%%%%%%%%%%%%%%%%%%%%%%%%%%%%%%%%%%%%%%%%%%%%%%%%%%%

%% P8: Talk about the broader impact and why our solution is the best to solve the challenges of mobile vision systems.

\vspace{0mm}
\noindent
\textbf{Summary of Contributions}.
%{\sysname} is particularly attractive to mobile vision systems. % for multiple reasons. 
%Mobile systems change context frequently, and thus need to run multiple DNNs to tackle different tasks. 
%Modern mobile vision systems support concurrent applications that compete for shared resources. 
%The available resources are dynamic and always changing due to events such as starting new applications, closing existing applications, and application priority changes. 
%
To the best of our knowledge, {\sysname} represents the first framework that enables resource-aware multi-tenant on-device deep learning for mobile vision systems.
%In doing so, {\sysname} address the key limitations in existing approaches for the emerging on-device
It contributes novel techniques that address the limitations in existing approaches as well as the unique challenges in continuous mobile vision.
We believe our work represents a significant step towards turning the envisioned continuous mobile vision into reality \cite{bahl2012vision, likamwa2015starfish, Naderiparizi2017}.
%We plan to release our framework and related tools in the near future.

%\input{challenges}
%!TEX root = mobicom2018.tex

%%
\section{ {\systitle} Overview}
\label{sec.overview}

Figure~\ref{dia.sys1} illustrates the architecture of {\sysname}, which is split into an \textit{offline stage} and an \textit{online stage}.

%%
%The offline stage consists of three phases: 1) DNN Pruning (\S xx), 2) DNN Recovery (\S xx), and 3) DNN Profiling (\S xx). 
The offline stage consists of three phases: model pruning (\S 3.1), model recovery (\S 3.2), and model profiling.

In the model pruning phase, {\sysname} employs a state-of-the-art \textit{Triplet Response Residual} (TRR) approach to rank the filters in a given deep learning model (i.e., \textit{vanilla model}) based on their importance and prune the filters iteratively.
%By following the ranking, filters in the model is iteratively pruned. 
%
In each iteration, less important filters are pruned, and the pruned model is retrained to compensate the accuracy degradation (if there is any) caused by filter pruning. 
The iteration ends when the pruned model could not meet the minimum accuracy goal set by the user.
%The iteration ends when the pruned model could not meet the minimum accuracy requirement set by the user. 
%
The smallest pruned model is called \textit{seed model}. 
As a result, a \textit{filter pruning roadmap} is created where each footprint on the roadmap is a pruned model with its filter pruning record. 
%This filter pruning roadmap is passed to the model recovery phase.
%Subsequently, based on \textit{filter importance ranking}, less important filters of the vanilla model are identified and \textit{pruned}. 
%The model then regains the accuracy by \textit{model re-training}.
%This three-step process is iterated until a \textit{seed model} is obtained that reaches the performance bottom line (i.e., lowest accuracy accepted by the user). 
%Meanwhile, {\sysname} records the filter pruning roadmap of pruned model generated by each three-step process. 

%%
In the model recovery phase, {\sysname} employs a novel \textit{model freezing and filter growing} (i.e., \textit{freeze-\&-grow}) approach to generate the multi-capacity model in an iterative manner.
%By following the filter pruning roadmap and the freeze-\&-grow approach, the \textit{multi-capacity model} of the vanilla model is iteratively generated.
%
Model recovery uses the seed model as the starting point.
In each iteration, \textit{model freezing} is first applied to freeze the parameters of all the model's filters.
%the filter parameters of the model is frozen.
%\textit{model freezing} is first applied to freeze
By following the filter pruning roadmap in the \textit{reverse order}, \textit{filter growing} is then applied to add the pruned filters back.
As such, a \textit{descendant model} with a larger capacity is generated and its accuracy is regained via retraining.
%As such, the final descen- dant model has the capacities of all the previous descendant models and is thus named multi-capacity model.
By repeating the iteration, a new descendant model is grown upon the previous one.
Thus, the final descendant model has the capacities of all the previous ones and is thus named multi-capacity model.

\begin{table}[t]
\centering
\scalebox{0.69}{
\begin{tabular}{|c|p{7.8cm}|}
\hline
\textbf{Terminology} 		& \textbf{\hspace{30mm} Explanation}                                                                                                                            \\ \hline
\textbf{Vanilla Model}              		& Off-the-shelf deep learning model (e.g., ResNet) trained on a given dataset (e.g., ImageNet). %It can achieve state-of-the-art performance.      
\\ \hline
\textbf{Pruned Model}               		& Intermediate result obtained in model pruning stage. %It contains only the most important filters.       
\\ \hline
\textbf{Seed Model}                 		& The smallest pruned model generated in model pruning which meets the minimum accuracy goal set by the user. It is also the starting point of model recovery stage.          \\ \hline
\textbf{Descendant Model}           	& A model grown upon the seed model in model recovery stage. It has a unique resource-accuracy trade-off. \\ \hline
\textbf{Multi-Capacity Model}       	& The final descendant model that has the capacities of all the previously generated descendant models.         \\ \hline
% but nested as a single model
\end{tabular}
}
\vspace{0mm}
\caption{Terminologies involved in {\sysname}.}
\label{tab.term}
\vspace{-12mm}
\end{table}

%{\sysname} first keeps the seed model unmodified by \textit{model freezing}. 
%It then patches certain amount of filters to the frozen seed model according to the recorded filter pruning paths. 
%Lastly, through \textit{recovery training}, it obtains a new model (\textit{descendant model}) with a larger inference capacity while the seed model remains intact. 
%This three-step process is iterated until it recovers the performance of the vanilla model. 
%Such model, with multiple descendant models with different inference capability, is named \textit{multi-capacity model}. 

%%
In the model profiling phase, given the specs of a mobile vision system, a profile is generated for each multi-capacity model including the inference accuracy, memory footprint, and processing latency of each of its descendent models.
%the system performance of the generated multi-capacity model is profiled through \textit{latency}, \textit{memory} and \textit{accuracy profiler} and send the model profile (i.e., \textit{L}, \textit{mem}, \textit{A}) to the mobile vision systems.

%%
%Finally, in the online stage, the {\sysname} scheduler (\S 4.3) continuously monitors changes in available runtime resources and application queries, and is triggered once a change is detected.
Finally, in the online stage, the resource-aware runtime scheduler (\S 3.3) continuously monitors events that change runtime resources.
%including changes in available runtime resources, starting new applications, closing existing applications, and changes in application priority, and is triggered once such event is occurred.
%The scheduler is triggered when a change is detected.
%when system runtime resources are changed, new applications are created, or currently running applications exit.
%
Once such event is detected, the scheduler checks up the profiles of all the concurrently running applications, selects the optimal descendant model for each application, and allocates the optimal amount of runtime resources to each selected descendant model to jointly maximize the overall inference accuracy and minimize the overall processing latency of all those applications.

%A monitored change will lead to two decisions by scheduler: \textit{optimal resource allocation} and \textit{optimal model selection}, such that the overall system performance is optimized. 
%It is achieved by encoding model profile, system runtime resources and application query into a cost function and then either minimizing the maximum cost or the sum of cost.

%%
For clarification purpose, Table~\ref{tab.term} summarizes the terminologies defined in this work and their brief explanations. 
%In the next section, we describe {\sysname} in detail.

% 4 pages
%&latex 
%!TEX root = mobicom2018.tex

\section{Design of {\systitle}}
\label{sec.algorithm}
\vspace{1mm}

%%
%\subsection{Filter Pruning for CNN Compression}
\subsection{Filter based Model Pruning}
\vspace{1mm}
%%
%\textit{4.1.1 Background on CNN Architecture} 
%\vspace{2mm}

\subsubsection{Background on CNN Architecture} 
$\\$ 
Before delving deep into filter pruning, it is important to understand the architecture of a convolutional neural network (CNN).
In general, a CNN consists of four types of layers: convolutional layers, activation layers, pooling layers, and fully-connected layers.
Due to the computational intensity of convolution operations, convolutional layers are the most computational intensive layers among the four types of layers.
Specifically, each convolutional layer is composed of a set of \textit{3D filters}, which plays the role of ``feature extractors''.
By convolving an image with these 3D filters, it generates a set of features organized in the form of \textit{feature maps}, which are further sent to the following convolutional layers for further feature extraction.

%%%%%%%%%%%%%%%%%%%%%%%%%%%%%%%%%%%%%%%%%%%%%%%%%%%%%%%%%%%%%%%%%

%%
\vspace{-1mm}
\subsubsection{Benefits of Filter Pruning}
\begin{figure}[t]
\centering
\hspace{-1.5mm}
\includegraphics[scale=0.60]{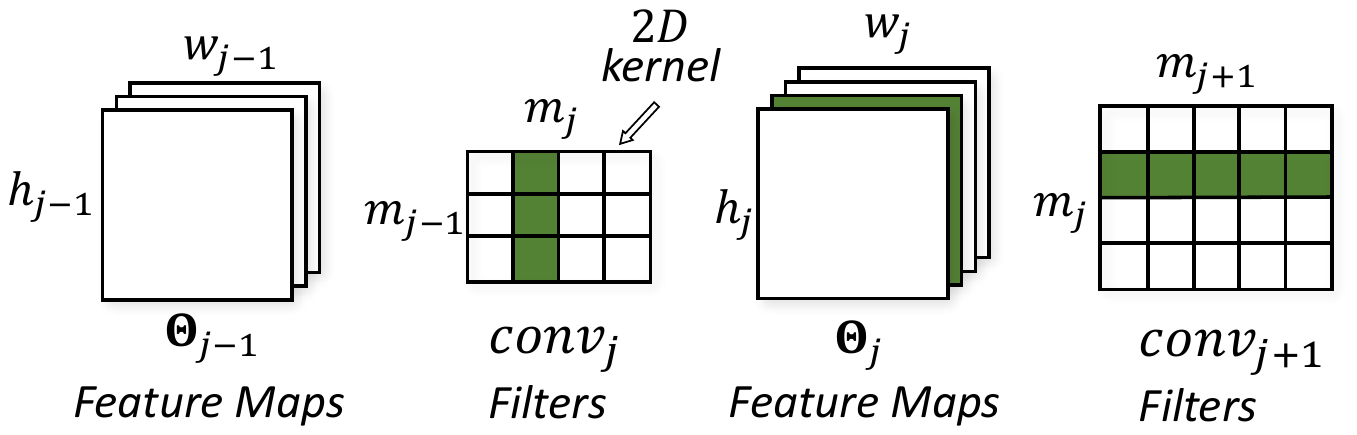}
\vspace{-2mm}
\caption{Illustration of filter pruning~\cite{li2016pruning}. By pruning filters, both model size and computational cost are reduced.}
\vspace{-3mm}
\label{dia.filter}
\end{figure}

$\\$ 
%
%% P1: 
%% What to talk about: Introduce the basics.
%
%Figure~\ref{dia.filter} illustrates the effect of filter pruning on reducing the size as well as the computational cost of a CNN model.
Figure~\ref{dia.filter} illustrates the details of filter pruning.
%pruning filters across two adjacent convolutional layers. 
%Pruning filters in a convolutional layer has effect on both current and its subsequent layer.
%
Let $\Theta_{j-1} \in \mathbb{R}^{w_{j - 1} \times h_{j - 1} \times m_{j - 1}}$ denote the input feature maps of the $j$th convolutional layer $conv_{j}$ of a CNN, where $w_{j - 1}$ and $h_{j - 1}$ are the width and height of each of the input feature maps; and $m_{j - 1}$ is the total number of the input feature maps. 
The convolutional layer $conv_{j}$ consists of $m_{j}$ 3D filters with size $k \times k \times m_{j-1}$ ($k \times k$ is the 2D kernel).
It applies these filters onto the input feature maps $\Theta_{j-1}$ to generate the output feature maps $\Theta_{j} \in \mathbb{R}^{w_{j} \times h_{j} \times m_{j}}$, where one 3D filter generates one output feature map. 
This process involves a total of $m_{j}k^{2}m_{j-1}w_{j}h_{j}$ floating point operations (i.e., FLOPs).

%% P2: 
%% What to talk about: Explain the benefit of filter pruning. 
%% Point that this paragraph makes: By pruning filters, both model size (i.e., parameters) and computational cost (i.e., FLOP) are reduced.
%
Since one 3D filter generates one output feature map, pruning one 3D filter in $conv_{j}$ (marked in green in $conv_{j}$) results in removing one output feature map in $\Theta_{j}$ (marked in green in $\Theta_{j}$), which leads to $k^2m_{j-1}$ parameter and $k^2m_{j-1}w_{j}h_{j}$ FLOPs reduction. 
Subsequently, $m_{j+1}$ 2D kernels applied onto that removed output feature map in the convolutional layer $conv_{j+1}$ (marked in green in $conv_{j+1}$) are also removed.
This leads to an additional $k^2m_{j+1}$ parameter and $k^2m_{j+1}w_{j+1}h_{j+1}$ FLOPs reduction. 
Therefore, by pruning filters, both model size (i.e., model parameters) and computational cost (i.e., FLOPs) are reduced~\cite{li2016pruning}. 
\vspace{-1mm}
\subsubsection{Filter Importance Ranking}
$\\$ 
%
%% P1:
%% What to talk about: Explain the goal. 
%
%After introducing how to prune filters in a layer, we then introduce the criteria in selecting which filters to be pruned.
The key to filter pruning is identifying less important filters.
%By pruning those filters, we can effectively reduce the memory footprint and computational cost of a DNN model while preserving its accuracy.
By pruning those filters, the size and computational cost of a CNN model can be effectively reduced.
% while preserving its accuracy.
%While pruning filters in convolutional layer effectively reduces the FLOP and hence reduces computational cost of a CNN model, pruning important filters may cause performance degradation of the neural network. Therefore, to ensure a minimum loss of performance of the neural network, we prune unimportant filters while preserving important ones.
%\xiao{The remain content of this paragraph can be put to (5.1.1background on pruning)}
%To rank the importance of filters in a single convolutional layer, state-of-the-art filter pruning techniques use $\mathcal{L}1$-norm or $\mathcal{L}2$-norm of a filter as an indicator of the importance of the filter~\cite{li2016pruning}. A filter with large norm is considered important because it generates a high magnitude filter response (i.e., feature map) which contributes more as the input of next layer. However, norm-based filter importance ranking may suffer in threefold: 1) it presents a local feature impact which does not necessarily lead to benign impact on the final prediction. In other words, it does not provide a global view of the network. So, it only provides a local optimal importance ranking at best; 2) it does not necessarily differentiate input with different classes; and 3) the filters with a large norm potentially cause model over-fitting.

%% P2:
%% What to talk about: Explain our method.
%
To this end, we propose a filter importance ranking approach named \textit{Triplet Response Residual} (TRR) to measure the importance of filters and rank filters based on their relative importance. 
%we employ a triplet loss function (i.e., triplet loss) to address this problem. Triplet loss has been successfully used for addressing variations in expression, pose, and illumination in the problem of unconstrained face recognition [24].
%
%
%Our TRR approach is inspired by two key intuitions. 
%Since a filter plays the role of a ``feature extractor'', the first key intuition behind our TRR approach is that \textit{a filter is important if it is able to extract features that are useful to differentiate images belonging to different classes}.
%Furthermore, in a CNN, features are organized in the form of feature maps. The second key intuition behind our TRR approach is that \textit{images from the same class would have more similar feature maps than images from different classes}.
%
Our TRR approach is inspired by one \textit{key intuition}: 
since a filter plays the role of ``feature extractor'', \textit{a filter is important if it is able to extract feature maps that are useful to differentiate images belonging to different classes}.
In other words, a filter is important if the feature maps it extracts from images belonging to the same class are more similar than the ones extracted from images belonging to different classes.
%We know that each filter generates a filter response for each input image. For images of the same class, the filter responses should be similar. For images of different classes, they should be different.

%%
Let \{$anc$, $pos$, $neg$\} denote a triplet that consists of an anchor image ($anc$), a positive image ($pos$), and a negative image ($neg$) where the anchor image and the positive image are from the same class, while the negative image is from a different class. 
By following the key intuition, TRR of filter $i$ is defined as:
\begin{multline}
TRR_i = \sum (\lVert \mathcal{F}_i(anc) - \mathcal{F}_i(neg) \rVert^{2}_{2}  -   \lVert \mathcal{F}_i(anc) - \mathcal{F}_i(pos) \rVert^{2}_{2}) 
%, \;\; i \in [1, m_{j}]
\end{multline} 
where $\mathcal{F}(\cdot)$ denotes the generated feature map. 
Essentially, TRR calculates the $\mathcal{L}2$ distances of feature maps between ($anc$, $neg$) and between ($anc$, $pos$), and measures the residual between the two distances. 
By summing up the residuals of all the triplets from the training dataset, the value of TRR of a particular filter reflects its capability of differentiating images belonging to different classes, acting as a measure of importance of the filter within the CNN model. 
%{\color{cyan}Moreover, filter responses of the triplet are calculated by feed forwarding input through all previous layers and hence have a global view of the network.}\xiao{ any rationale behind this?}
%As such, the larger TRR is, the more important the filter is to the entire network. 
%By pruning the unimportant filters of each layer, we are able to find a global optimal pruned model. 
%At the core of our TRR approach is the utilization of a triplet loss function (i.e., triplet loss)~\cite{schroff2015facenet}.
% that measures the  
%TRR ranking is inspired by the triplet loss of representation learning~\cite{schroff2015facenet}. 
%While triplet loss minimizes the distance between anchor ($anc$) and positive ($pos$) and maximizes the distance between anchor and negative ($neg$), TRR ranks the ability of filters to minimize and maximize those two distances, respectively.

%%%%%%%%%%%%%%%%%%%%%%%%%%%%%%%%%%%%%%%%%%%%%%%%%%%%%%%%%%%%%%%%%

%%
\subsubsection{Performance of Filter Importance Ranking}
\begin{figure}[t]
\centering
\hspace{-1mm}
\includegraphics[scale=0.38]{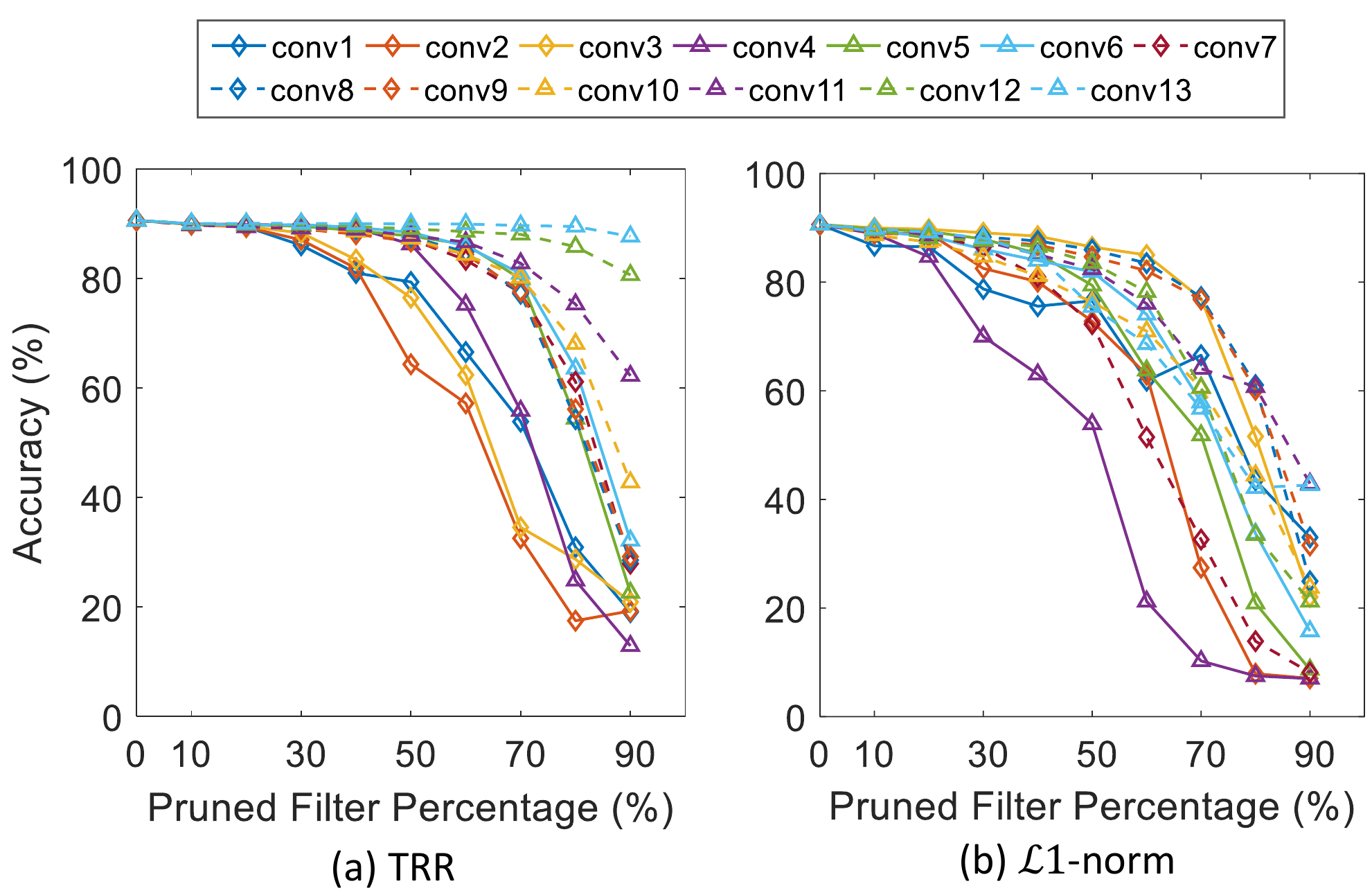}
\vspace{-4mm}
\caption{Filter importance profiling performance of (a) TRR and (b) $\mathcal{L}1$-norm on VGG-16 trained on CIFAR-10.}
\vspace{-2mm}
\label{dia.trr}
\end{figure}

%\subsubsection{Pruning Filters Across Convolutional Layers}
%
$\\$ 
%
%% P1: 
%% What to talk about: Explain the experiment.
%% P2: 
%% What to talk about: Explain the performance of our method.
%% Point that this paragraph makes: Our approach is effective at identifying less important filters.
%
Figure~\ref{dia.trr}(a) illustrates the filter importance profiling performance of our TRR approach on VGG-16~\cite{simonyan2014very} trained on the CIFAR-10 dataset~\cite{krizhevsky2009learning}.
%We use the vanilla {\vgg} trained on CIFAR-10 as a representative of other CNN models and datasets. 
%
The vanilla VGG-16 model contains 13 convolutional layers.
Each of the 13 curves in the figure depicts the top-1 accuracies when filters of one particular convolutional layer are pruned while the other convolutional layers remain unmodified. 
Each marker on the curve corresponds to the top-1 accuracy when a particular percentage of filters is pruned. 
As an example, the topmost curve (blue dotted line with blue triangle markers) shows the accuracies are 89.75\%, 89.72\% and 87.40\% when 0\% (i.e., vanilla model), 50\% and 90\% of the filters in the 13th convolutional layer $conv_{13}$ are pruned, respectively. 

We have two key observations from the filter importance profiling result. 
First, we observe that our TRR approach is able to effectively identify redundant filters within each convolutional layer. 
In particular, the accuracy remains the same when 59.96\% of the filters in $conv_{13}$ are pruned.
This indicates that these pruned filters, identified by TRR, are redundant.
By pruning these redundant filters, the vanilla VGG-16 model can be effectively compressed without any accuracy degradation.
%This indicates that  XXX\% of the filters are identified to be TRR is able to 
%across all 13 convolutional layers, TRR is able to identify the accuracy does not drop when as much as 20\% of filters of one particular layer are pruned.
%for each of 13 convolutional layers, the accuracy does not drop when certain percentage of filters identified by TRR are pruned. 
%For example, , indicating that these pruned filters, identified by TRR, are redundant and thus have no contributions to accuracy.
% All convolutional layers have redundant filters.
%
Second, we observe that our TRR approach is able to effectively identify convolutional layers that are more sensitive to filter pruning. 
This is reflected by the differences in accuracy drops when the same percentage of filters are pruned at different convolutional layers. 
This sensitivity difference across convolutional layers has been taken into account in the iterative filter pruning process.
%For example, when 50\% and 90\% of unimportant filters are pruned the accuracies of $conv_{13}$ and $conv_{2}$ (marked as red line with diamond markers) drop from 89.75\% to 89.72\% and 87.40\% , and to 64.33\% and 19.28\%, respectively. 
%This indicates that $conv_{13}$ is much less sensitive to filter pruning compared to $conv_{2}$, which indicates that $conv_{13}$ contains more unimportant filters. 

%% P3: 
%% What to talk about: Compare with state-of-the-art method. 
%% Point that this paragraph makes: Our approach is better.
%
To demonstrate the superiority of our TRR approach, we have compared it with the state-of-the-art filter pruning approach.
The state-of-the-art filter pruning approach uses $\mathcal{L}1$-norm of a filter to measure its importance~\cite{li2016pruning}.
%To make a fair comparison, we repeat the filter importance profiling on the same vanilla VGG-16 model trained on CIFAR-10 except that $\mathcal{L}1$-norm is used as the measure of filter importance. 
%
Figure~\ref{dia.trr}(b) illustrates the filter importance profiling performance of $\mathcal{L}1$-norm on the same vanilla VGG-16 model trained on the CIFAR-10 dataset.
By comparing Figure~\ref{dia.trr}(a) to Figure~\ref{dia.trr}(b), we observe that TRR achieves better accuracy than $\mathcal{L}1$-norm at almost every pruned filter percentage across all 13 curves.
%for each curve, 
%As shown, the accuracy of this model is sacrificed more severely when it is pruned by $\mathcal{L}1$-norm than TRR.
%
As a concrete example, TRR achieves an accuracy of 89.72\% and 87.40\% when 50\% and 90\% of the filters at $conv_{13}$ are pruned respectively, while $\mathcal{L}1$-norm only achieves an accuracy of 75.45\% and 42.65\% correspondingly.
%Overall, TRR outperforms $\mathcal{L}1$-norm by 8.6\% accuracy across all pruned filter percentages of 13 convolutional layers. 
%
This result indicates that the filters pruned by TRR have much less impact on accuracy than the ones pruned by $\mathcal{L}1$-norm, demonstrating that TRR outperforms $\mathcal{L}1$-norm at identifying less important filters.

%%%%%%%%%%%%%%%%%%%%%%%%%%%%%%%%%%%%%%%%%%%%%%%%%%%%%%%%%%%%%%%%%

%%
\subsubsection{Filter Pruning Roadmap}
$\\$ 
By following the filter importance ranking provided by TRR, we iteratively prune the filters in a CNN model.
%and the pruned model is retrained to regain accuracy.
During each iteration, less important filters across convolutional layers are pruned, and the pruned model is retrained to compensate the accuracy degradation (if there is any) caused by filter pruning.
The iteration ends when the pruned model could not meet the minimum accuracy goal set by the user. 
%As a result, a \textit{filter pruning roadmap} is created where each footprint on the roadmap is a pruned model with a unique resource-accuracy profile along with the pruned filters. 
As a result, a \textit{filter pruning roadmap} is created where each footprint on the roadmap is a pruned model with its filter pruning record. %and its unique resource-accuracy profile. 
The smallest pruned model on the roadmap is called \textit{seed model}.
This filter pruning roadmap is used to guide the model recovery process described below.
%in the next subsection.

%\squishlist{
%\item{Given a trained model, profile TRR of all filters of all convolutional layers across the entire CNN model.} 
%\item{Prune the filters that are less important according to TRR ranking of each layer and prune the number of filters across the network according to susceptibility indicates by TRR profile.}
%\item{Retrain the pruned CNN model until performance converges.}
%\item{Iterate the previous three steps until the pruned CNN model meets the system resource bottom lines (e.g., FLOP, FPS or runtime memory).}
%\item{The pruning footprint (i.e., pruned filters, model architectures) forms a pruning path, which will be used to recover the model in the second stage.}
%}\squishend
%To sum up, the iterative filter pruning ends up with: 1) an efficient seed model that meets the minimum requirement of a mobile device; 2) a series of trained intermediate models with an optimal filter configuration; and 3) the most effective filter \textit{pruning path} to reach the seed model.

%%%%%%%%%%%%%%%%%%%%%%%%%%%%%%%%%%%%%%%%%%%%%%%%%%%%%%%%%%%%%%%%%

%%
%\subsection{Freeze-\&-Grow for DNN Recovery}
\subsection{Freeze-\&-Grow based Model Recovery}

\subsubsection{Motivation and Key Idea}
$\\$ 
%
%% Talk about the key drawback of the pruned models generated by filter pruning. Why we could not use them for scheduling.
%
The filter pruning process generates a series of pruned models, each of which acting as a model variant of the vanilla model with a unique resource-accuracy trade-off.
However, due to the retraining step within each pruning iteration, these pruned models have different model parameters, and thus are independent of each other.
%do not share parameters with each other.
% 
Therefore, although these pruned models provide different resource-accuracy trade-offs, keeping all of them locally in resource-limited mobile systems is practically infeasible.

%Thus far, we have obtain a series of independent models with different capacities (inference accuracy and latency).
%As mentioned in \S 2, keeping all model variants locally or downloading them from the cloud is impractical and the overhead of switching between them is not negligible.

%%
To address this problem, we propose to generate a single \textit{multi-capacity model} that acts equivalently as the series of pruned models to provide various resource-accuracy trade-offs but has a model size that is much smaller than the accumulated model size of all the pruned models. 
%%we propose to train a multi-capacity model that acts equivalently as these series independent pruned models while only using memory of a single model size. 
%Such a multi-capacity model can also greatly reduce the overhead incurred by model switching.
%
This is achieved by an innovative \textit{model freezing and filter growing} (i.e., \textit{freeze-\&-grow}) approach.

In the remainder of this section, we describe the details of the freeze-\&-grow approach and how the multi-capacity model is iteratively generated.
%In the remaining of this subsection, we describe how the multi-capacity model is iteratively generated based on the freeze-\&-grow approach.

%While the pruned model has reduced size and FLOP, it cannot regain its accuracy once the sacrifice has been made and hence is impossible to meet all runtime resource requirements.
%%
%To this end, we propose a freeze-and-grow approach that generates a multi-capacity model that can regain its sacrificed accuracy as demand.
%has an optimized model in terms of efficiency, it suffers a graceful drop of accuracy and only offers a fixed trade-off between model performance and inference speed, and hence does not provide enough flexibility to meet various requirements in a mobile environment. To this end, we propose DNN recovery, a novel training regime that recovers performance from the seed model and generate a series of descendant models with different trade-offs between performance and inference speed. More importantly, to minimize the overhead of model switching, these descendant models are designed to share parameters with another, forming a nested DNN. 

%%%%%%%%%%%%%%%%%%%%%%%%%%%%%%%%%%%%%%%%%%%%%%%%%%%%%%%%%%%%%%%%%

%%
\subsubsection{Model Freezing and Filter Growing}
$\\$ 
%
%The seed model suffers drop of accuracy because its convolutional layers lack enough learning capability due to a limited number of filters. To compensate for the drop of accuracy, we add the number of filters of each convolutional layer. 
%Although the seed model obtained via DNN pruning has an optimized model in terms of efficiency, it suffers a graceful drop of accuracy and only offers a fixed trade-off between model performance and inference speed, and hence does not provide enough flexibility to meet various requirements in a mobile environment.
%% P1:
% Talk about the idea of Model Freezing and Filter Growing; and why it solves the key drawback of the pruned models generated by filter pruning
%%%% recap what we have achieved so far, a multi-capacity model
%%%%
%%
The generation of the multi-capacity model starts from the seed model derived from the filter pruning process.
By following the filter pruning roadmap and the freeze-\&-grow approach, the multi-capacity model is iteratively created.
%Importantly, in order to regain certain capability with least possible additional filters, we add filters by exactly back tracking \textit{filter pruning roadmap} obtained from filter pruning.  
%%
%process starts from the seed model and iteratively recovers to each pruned model following the pruning roadmap, which can be viewed as the reverse of DNN pruning, thus we call it DNN recovery.
%To this end, we propose a novel training regime that recovers performance from the seed model and generate a series of descendant models with different trade-offs between performance and inference speed. 
%Following the filter pruning roadmap in an opposite direction, we iteratively recover the pruned model.

%% P2:
% Talk about the details of Model Freezing and Filter Growing.
%Figure~\ref{dia.grow} shows the effect of filter growing between two adjacent convolutional layers. 
%As shown, by adding an additional 3D filter to $conv_{j}$ filters (marked as green), one additional feature map is created in $\Theta_{j}$ feature maps (marked as green). 
%Meanwhile, to account for the increased feature map in $\Theta_{j}$, $m_{j+1}$ 2D filters are added to $conv_{j+1}$ filters (marked as green). 
%
%Moreover, by adding an additional 3D filters to $conv_{j+1}$ filters (marked as orange), one additional feature map is created in $\Theta_{j+1}$ feature maps (marked as orange).
%As such, by adding one 3D filter and $m_{j+1}$ 2D filters in two consecutive convolutional layers respectively, one additional feature maps could be generated. 
%This leads to an increased learning capability of this model.

\begin{figure}[t]
\centering
\hspace{-1.5mm}
\includegraphics[scale=0.71]{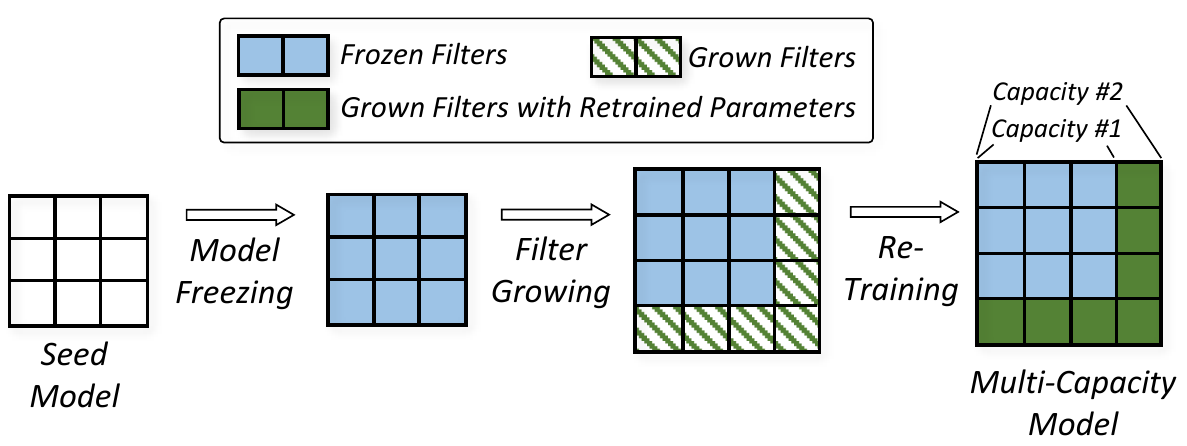}
\vspace{-6mm}
\caption{Illustration of model freezing and filter growing.}
%For illustration purpose, only one convolutional layer is depicted.
\vspace{-4mm}
\label{dia.grow}
\end{figure} 

Figure~\ref{dia.grow} illustrates the details of model freezing and filter growing during the first iteration. %model freezing and filter growing.
For illustration purpose, only one convolutional layer is depicted. 
%to represent the entire model and show the process of training a two-capacity model.
%
As shown, given the seed model, we first apply \textit{model freezing} to freeze the parameters of all its filters (marked as blue squares). 
Next, since each footprint on the roadmap has its filter pruning record, we follow the filter pruning roadmap in the \textit{reverse order} and apply \textit{filter growing} to add the pruned filters back (marked as green stripe squares). 
%Grow filters of each convolutional layer cross the entire network, backtracking the \textit{pruning path} obtained from DNN pruning.
%
With the added filters, the capacity of this \textit{descendant model} is increased.
Lastly, we retrain this descendant model to regain accuracy.
%Re-train the grown model until the accuracy is recovered, and generate a recovered descendant model.
%
It is important to note that during retraining, since the seed model is frozen, its parameters are not changed; only the parameters of the added filters are changed (marked as green solid squares to indicate the parameters are changed).
%As shown, this three-step process starts with a seed model that all parameters are trainable. 
%Next, we freeze the seed model such that all parameters are untrainable (marked as blue).
%We then grow trainable filters (marked as green stripes) to the frozen seed model such that the capacity of this model is increased.
%Finally, this model regains accuracy by re-training the grown filters (marked as green).
%
As such, we have generated a single model that not only has the capacity of the seed model but also has the capacity of the descendant model. 
Moreover, the seed model \textit{shares} all its model parameters with the descendant model, making itself \textit{nested} inside the descendant model without taking extra memory space. 
%As a result, a single model which not only has an increased capability (capacity \#2) but also the capability of seed model (capability \#1) is generated. 

%%
By repeating the iteration, a new descendant model is grown upon the previous one. 
%This new descendant model is then fed into the next iteration.
% each descendant model is built upon its previous model(s) 
%As such, the final descendant model contains all the previous descendant models.
%
As such, the final descendant model has the capacities of all the previous descendant models and is thus named \textit{multi-capacity model}.
\vspace{-2mm}
\subsubsection{Superiority of Multi-Capacity Model}
$\\$ 
%The generated multi-capacity model has the following three key advantages.
The generated multi-capacity model has three advantages.

\vspace{1mm}
\noindent
\textbf{One Compact Model with Multiple Capabilities}.
The generated multi-capacity model is able to provide multiple capacities nested in a single model. 
This eliminates the need of installing potentially a large number of independent model variants with different capacities.
%It provides multiple capabilities equivalent to a series of descendent models, eliminating the need to carry a large amount of descendent models.
%
Moreover, by sharing parameters among descendant models, the multi-capacity model is able to save a large amount of memory space to significantly reduce its memory footprint.

%can be viewed as containing several descendant models with different trade-offs between performance and inference speed. 
%%%%% first benefit: reduce storage size
%Moreover, training a multi-capacity model can be viewed as training a single model to contain all its previous descendant models. 
%A resulting benefit is that we can use the multi-capacity model as a replacement for potentially numerous descendant models, saving huge amount of storage. 
%Moreover, the efficient model provides multiple inference capabilities encapsulated in a single neural network. This is accomplished by parameter sharing among descendant models.

%%
\vspace{1mm}
\noindent
\textbf{Optimized Resource-Accuracy Trade-offs}.
Each capacity provided by the multi-capacity model has a unique optimized resource-accuracy trade-off. 
Our TRR approach is able to provide state-of-the-art performance at identifying and pruning less important filters.
As a result, the multi-capacity model delivers state-of-the-art inference accuracy under a given resource budget.
%It provides state-of-the-art inference accuracies under various resource budgets due to sophisticated DNN pruning and recovery processes.

%%
\vspace{1mm}
\noindent
\textbf{Efficient Model Switching}.
%
%Figure~\ref{dia.nested} depicts the upgrade and downgrade mechanism of a nested DNN. 
%For simplicity, here we consider a DNN model with only one convolutional layer followed by a fully-connected layer that outputs inference results. 
%As shown, a nested DNN is able to switch from current descendant (in the middle) to a larger descendant model by growing filters (marked as green), or to a smaller descendant model by shedding filters (marked as gray). Essentially, the switch to a larger descendant model is a process of trade-in resource (e.g., memory, FLOPs and energy consumption) in exchange for a higher accuracy; the switch to a smaller descendant model is a process of sacrificing a graceful drop of accuracy in exchange for a faster and lighter inference model. 
%
%%%%% second benefit: reduce switching overhead, 
%%
Because of parameter sharing, the multi-capacity model is able to switch models with little overhead. %incurs minimum overhead during model switching.
Switching independent deep learning models causes significant overhead.
This is because it requires to page in and page out the \textit{entire} deep learning models.  
Multi-capacity model alleviates this problem in an elegant manner by only requiring to page in and page out a very small portion of deep learning models.

%Nested DNN alleviates the problem because it does not require to page-in the entire model, but only requires to page-in, from storage or cache, a small portion of weights that is not shared between the two descendant models. 
%Moreover, if the nested DNN is requested to switch to a smaller descendant model, it does not produce any overhead from memory page-in. %

\begin{figure}[t]
\centering
\hspace{-1.5mm}
\includegraphics[scale=0.75]{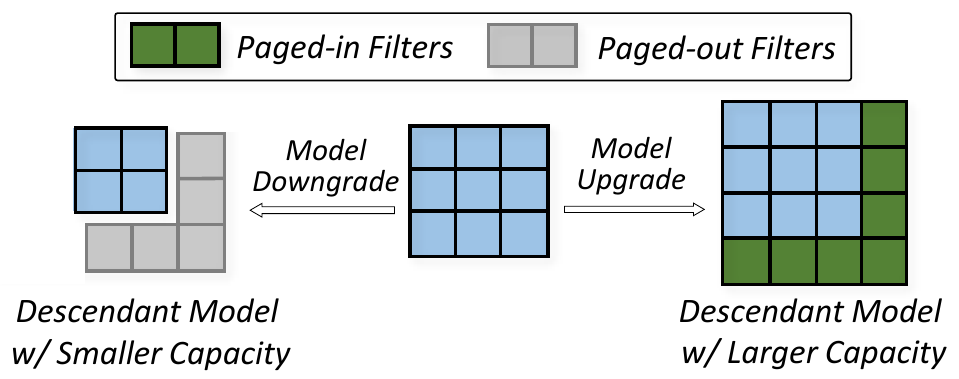}
\vspace{-1mm}
\caption{Illustration of model switching (model upgrade vs. model downgrade) of multi-capacity model.}
%For illustration purpose, only one convolutional layer is depicted.
\vspace{-2mm}
\label{dia.nested}
\end{figure}

%%
Figure~\ref{dia.nested} illustrates the details of model switching of multi-capacity model. 
%For simplicity, we consider a model with only one convolutional layer followed by a fully-connected layer that outputs inference results.
For illustration purpose, only one convolutional layer is depicted.
As shown, since each descendant model is grown upon its previous descendant models, when the multi-capacity model is switching to a descendant model with larger capability (i.e., \textit{model upgrade}), it incurs \textit{zero page-out overhead}, and only needs to page in the extra filters included in the descendant model with larger capability (marked as green squares).
When the multi-capacity model is switching to a descendant model with smaller capability (i.e., \textit{model downgrade}), it incurs \textit{zero page-in overhead}, and only needs to page out the filters that the descendant model with smaller capability does not have (marked as gray squares).
As a result, the multi-capacity model significantly reduces the overhead of model page in and page out, making model switching extremely efficient.
% by avoiding shared weights loading, making it suitable for performance trade-off.
%Therefore, the nested DNN is especially suitable for computing systems that dynamically change runtime system resource (e.g., memory or FLOPs).   

%As shown, a multi-capacity is able to switch from current descendant (in the middle) to a latter descendant model by growing filters (marked as green), or to a smaller descendant model by shedding filters (marked as gray).

%%

%it does not need to page out the entire model when switching to one of its previous models (downgrade). 
%Besides, when switching to a latter model (upgrade), it only needs to page in the extra weights that the latter model have grown.

%As a result, it is able to meet different system requirements and switch between each other with little overhead. 
%The switch process incurs a minimum overhead of memory read and write.
%The multi-capacity model can reduce more than half of the model switching overhead.
%Traditional model switching approaches page in and out entire DNN model and thus cause large overhead, especially when the models are of large size.
%

%%Traditionally, the process of switching between DNN models incurs a large overhead due to: 1) DNN models are typically heavyweight because it contains huge amount of weights and hence the memory of a mobile device cannot load or cache many of them simultaneously; and 2) the memory page-in and page-out of heavyweight DNN models cause large overhead. 
%

\subsection{Resource-Aware Scheduler}

\subsubsection{Motivation and Key Idea}
%\textbf{User Preferences}.
%
$\\$ 
The creation of the multi-capacity model enables {\sysname} to jointly maximize the performance of vision applications that are concurrently running on a mobile vision system.
%For a vision application, the two important metrics for evaluating its performance are inference accuracy and inference latency.
%In the meantime, different users have different desired preferences.
%enables unprecedented flexibility for a deep learning-based mobile vision application to achieve trade-off between accuracy and resource use.
%
This possibility comes from two \textit{key insights}.
%Mobile vision applications have diverse goals.
%This variety comes from the following two realities in mobile setting:
%
First, while a certain amount of runtime resources can be traded for an accuracy gain in some application, the same amount of runtime resources may be traded for a \textit{larger} accuracy gain in some other application.
Second, for applications that do not need real-time response and thus can tolerate a relatively large processing latency, we can \textit{reallocate} some runtime resources from those latency-tolerant applications to other applications that need more runtime resources to meet their real-time goals.
{\sysname} exploits these two key insights by encoding the inference accuracy and processing latency into a \textit{cost function} for each vision application, which serves as the foundation for resource-aware scheduling.

%%
%In mobile vision systems, users may expect different optimization effect from the scheduler.
%The difference in expectation has two-fold:
%\begin{enumerate}
	%\item 	
%Some of the APPs may prefer accuracy over FPS while the other may prefer FPS over accuracy.
%To satisfy this need, we provide a tunable factor $\alpha$ for users to determine the trade-off between accuracy and FPS for each APP. 
%A larger $\alpha$ will favor more efficient inferences, resulting in higher FPS and lower accuracy.
%A smaller $\alpha$ will favor higher accuracy, resulting in lower FPS.
	
%\item In some scenarios, users may prefer to fairly balance the runtime performance among all running APPs while in other scenarios, users may prefer to favor those APPs that can deliver higher performance.
	%To satisfy this need, we provide two scheduling schemes: MinTotalCost and MinMaxCost. Users can select corresponding scheme during scheduling.
%\end{enumerate}

%%%%%%%%%%%%%%%%%%%%%%%%%%%%%%%%%%%%%%%%%%%%%%%%%%%%%%%%%%%%%%%%%

%%
%\vspace{1mm}
%\noindent
%\textbf{Cost Function}.
\subsubsection{Cost Function}
$\\$ 
%
%To incorporate the user preference on inference accuracy and inference latency of a vision application running on a mobile vision system, {\sysname} encodes these two important metrics into a \textit{cost function} for each vision application.
%Specifically, 
Let $V$ denote the set of vision applications that are concurrently running on a mobile vision system, and let $A_{min}(v)$ and $L_{max}(v)$ denote the minimum inference accuracy and the maximum processing latency goals set by the user for application $v \in V$. 
%and let $v$ denote the vision application $v \in V$. 
%where $A_{min}(v)$ is its minimum accuracy as, maximum latency as $L_{max}(v)$. 
%
Additionally, let $M_{v}$ denote the multi-capacity model generated for application $v$, and let $m_v$ denote a descendant model $m_v \in M_{v}$.
% We define the cost of running model $m_v$ for APP $v$ with $u_{v}$ as:
The cost function of the descendant model $m_v$ for application $v$ is defined as follows:

\vspace{-1mm}
\begin{equation}
\begin{aligned}
C(m_v, u_v, v) = (A_{min}(v)-A(m_v)) + \\
\alpha \cdot \max(0,\frac{L(m_v)}{u_v} - L_{max}(v))
\end{aligned}
\end{equation}
where $A(m_v)$ is the inference accuracy of $m_v$,
$u_v \in (0,1] $ is the computing resource percentage allocated to $v$,  
and $L(m_v)$ is the processing latency of $m_v$ when 100\% computing resources are allocated to $v$. 
%chosen by the user preference.
%the profiled accuracy and latency for model $m_v$ 
%$I(m_v)$ represents the initialization latency for loading model $m_v$. If model has already been loaded, then $I(m_v) = 0$.
%%
%The value of $A(m_v)$ is obtained at the multi-capacity model development stage and the value of $L(m_v)$ is obtained via model profiling.
%

%%
Essentially, 
%the first term in the cost function represents the accuracy difference that is blow minimum accuracy.
%the first term in the cost function represents the penalty for selecting a descendant model $m_v$ that has an inference accuracy lower than the minimum accuracy goal.
the first term in the cost function promotes selecting the descendant model $m_v$ whose inference accuracy is as high as possible.
% set by the user.
%
The second term in the cost function penalizes selecting the descendant model $m_v$ that has a processing latency higher than the maximum processing latency goal.
Since the video input is streamed at a dedicated frame rate, there is no reward for achieving a processing latency lower than $L_{max}(v)$.
% set by the user.
%represents the latency difference that is above maximum latency. 
%Minimizing both terms can be viewed as selecting a descendant model $m_v$ and allocating some amount of runtime resources $u_v$ to ensure that the accuracy is above $A_{min}(v)$ and latency is below $L_{max}(v)$.
%$\{m_v|v \in V\}$
%In other words, 
%
$\alpha \in [0,1]$ is a knob set by the user to determine the latency-accuracy trade-off preference.
A large $\alpha$ weights more on the penalty for latency while a small $\alpha$ favors higher accuracy.
%efficient inferences, resulting in higher FPS and lower accuracy.
%A smaller $\alpha$ will favor higher accuracy, resulting in lower FPS.
%
%Our goal is to select an appropriate descendent model $m_v$ for the running application $v$, and to determine the amount of available runtime resources $u_v$ to allocate to $m_v$ such that the cost function is minimized. 
%runtime resources $\{u_v|v \in V\}$ for the running apps $V$ such that cost function is minimized without exceeding available memory and CPU resources.

%%%%%%%%%%%%%%%%%%%%%%%%%%%%%%%%%%%%%%%%%%%%%%%%%%%%%%%%%%%%%%%%%

%%
\subsubsection{Scheduling Schemes}
$\\$ 
Given the cost function of each descendant model of each concurrently running application, the resource-aware scheduler incorporates two widely used scheduling schemes to jointly maximize the performance of concurrent vision applications for two different optimization objectives.
%The scheduling problem is formulated as follows.
%Depending on users' need, we provide two scheduling schemes: \textit{MinTotalCost} and \textit{MinMaxCost}. 

%%
\vspace{1.5mm}
\noindent
\textbf{MinTotalCost}.
The MinTotalCost (i.e., minimize the total cost) scheduling scheme aims to minimize the total cost of all concurrent applications.
This optimization problem can be formulated as follows: 

\vspace{0mm}
\begin{equation}
\min_{u_v,m_v \in M_{v}} \sum \limits_{v \in V}C(m_v,u_v,v)
\end{equation}
\vspace{-1mm}
\begin{equation*}
s.t. \quad   \sum \limits_{v \in V} S(m_v) \le S_{max}, \quad \sum \limits_{v \in V} u_v \le 1
\vspace{2mm}
\end{equation*}
%\vspace{-1mm}
%\begin{equation*}
%\sum \limits_{v \in V} u_v \le 1
%\end{equation*}
%
where $S(m_v)$ denotes the runtime memory footprint of the descendant model $m_v$.
The total memory footprint of all the concurrent applications cannot exceed the maximum memory space of the mobile vision system denoted as $S_{max}$.

Under the MinTotalCost scheduling scheme, the resource-aware scheduler favors applications with lower costs and thus is optimized to allocate more runtime resources to them.
%This scheduling scheme is particularly useful to mobile vision systems XXX.
%applications with higher performance will be given higher priority with more resources while those APPs with lower performance may be 'sacrificed'.

%In addition to cost, we also need to consider memory constraint $S_{max}$ and computing resource constraint. We have the following constraints:
%\vspace{-2mm}
%\begin{equation}
%	\sum \limits_{v \in V} S(m_v) \le S_{max}
%\end{equation}
%\vspace{-3mm}
%\begin{equation}
%	\sum \limits_{v \in V} u_v \le 1
%\end{equation}
%\noindent
%where $S(m_v)$ denotes the runtime memory of model $m_v$. 
%\vspace{1mm}

%%%%%%%%%%%%%%%%%%%%%%%%%%%%%%%%%%%%%%%%%%%%%%%%%%%%%%%%%%%%%%%%%

%%
\vspace{1.5mm}
\noindent
\textbf{MinMaxCost}.
The MinMaxCost (i.e., minimize the maximum cost) scheduling scheme aims to minimize the cost of the application that has the highest cost.
This optimization problem can be formulated as follows: 

\vspace{-1mm}
\begin{equation}
\min_{u_v,m_v \in M_{v}} k
\end{equation}
\vspace{-1mm}
%\begin{equation*}
%s.t. \quad   \forall v: C(m_v,u_v,v) \le k
%\end{equation*}
%\vspace{-1mm}
\begin{equation*}
s.t. \quad   \forall v: C(m_v,u_v,v) \le k, 
\vspace{2mm}
\end{equation*}
\begin{equation*}
\sum \limits_{v \in V} S(m_v) \le S_{max}, \quad \sum \limits_{v \in V} u_v \le 1
\vspace{2mm}
\end{equation*}
where the cost of any of the concurrently running applications must be smaller than $k$ where $k$ is minimized.

Under the MinMaxCost scheduling scheme, the resource-aware scheduler is optimized to fairly allocate runtime resources to all the concurrent applications to balance their performance.

%\noindent
%The optimization results are called optimal scheme.
%These two optimization problems represent two scheduling scenarios. In later experiments we can see that NestDNN outperforms baseline in both scenarios.

%%
%With all the model profiles given by profiler, the scheduler is responsible for determining the optimal running scheme.
%The scheduler takes the following as input:
%1) model performance profile for each running APP in system, given by the profiler;
%2) a weight factor controlling the preference between accuracy and latency;
%3) minimum inference accuracy and maximum inference latency.
%It outputs an optimal running scheme, which is comprised of a selected descendant model and allocated computing resources for each APP.

%%%%%%%%%%%%%%%%%%%%%%%%%%%%%%%%%%%%%%%%%%%%%%%%%%%%%%%%%%%%%%%%%

%%
\subsubsection{Cached Greedy Heuristic Approximation}
$\\$ 
%
%The next problem is how to solve the scheduling problem.
%
%
Solving the nonlinear optimization problems involved in MinTotalCost and MinMaxCost scheduling schemes is computationally hard.
To enable real-time online scheduling in mobile systems, we utilize a greedy heuristic inspired by \cite{zhang2017live} to obtain approximate solutions.
% of the optimization problems. %approximate the optimal scheme.
%The procedure of greedy method is described as follows. 

%%
Specifically, we define a minimum indivisible runtime resource unit $\Delta u$ (e.g., $1\%$ of the total computing resources in a mobile vision system) and start allocating the computing resources from scratch. % (no CPU resource has been allocated yet).
For MinTotalCost, 
we allocate $\Delta u$ to the descendent model $m_v$ of application $v$ such that $C(m_v, \Delta u, v) $ has the smallest cost increase among other concurrent applications.
For MinMaxCost, 
we select application $v$ with the highest cost $C(m_v,u_v,v)$, and allocate $\Delta u$ to $v$ and choose the optimal descendent model $m_v = \arg \min_{m_v} C(m_v,u_v,v)$ for $v$.
For both MinTotalCost and MinMaxCost, the runtime resources are iteratively allocated until exhausted.
% or the cost stops increasing.

%Although the greedy heuristic can approximate the optimal solution and be solved within deterministic time, it is not always feasible in mobile systems.
%The reason is that computing resource is limited in mobile systems, especially when multiple applications are running. 
%In this case, the heuristic method cannot solve the optimization problem in time.
%
The runtime of executing the greedy heuristic can be further shortened via the caching technique.
This is particularly attractive to mobile systems with very limited resources.
%Therefore, we propose an cached greedy heuristic approximation, which is aimed for mobile setting.
%The idea of cached greedy heuristic is that 
Specifically, when allocating the computing resources, instead of starting from scratch, we start from the point where a certain amount of computing resources has already been allocated.
%we can solve the problem based on the last optimal scheduling result.
%
For example, we can cache the unfinished running scheme where $70\%$ of the computing resources have been allocated during optimization.
In the next optimization iteration, we directly start from the unfinished running scheme and allocate the remaining $30\%$ computing resources, thus saving $70\%$ of the optimization time.
%In extreme cases, we can start from the point where $99\%$ CPU resources have been allocated, thus saving $99\%$ of the optimization time.
%
To prevent from falling into a local minimum over time, a complete execution of the greedy heuristic is performed periodically to enforce the cached solution to be close to the optimal one.

%%\subsubsection{Triggering Scheduler}
%%
%The last problem is to decide when the scheduler should be triggered. 
%In fact, once the scheduler outputs an optimal scheme, this optimal scheme will not change unless: 
%\begin{enumerate}
%	\item \textit{External Variation}: the available resources that can be allocated to all APPs is changed;
%	\item \textit{Internal Variation}: new APPs are created or some of the running APPs are closed.
%\end{enumerate}
%The scheduling process will be triggered when either of these two cases occurs. This strategy is more flexible and efficient thus suitable for mobile systems.
%Specifically, to handle external variation, we monitor two crucial system performance metrics. The first one is memory variation $\Delta mem$. The second one is CPU utilization variation $\Delta cpu$.
%We implement a system performance monitor and when the monitor observes a notable change in $\Delta mem$ or $\Delta cpu$, the scheduler will be triggered.
%In other words, the scheduler will be activated when $\Delta mem \ge mem_{th}$ or $\Delta cpu \ge cpu_{th}$. $mem_{th}$ and $cpu_{th}$ are pre-determined parameters.
%To handle internal variation, an APP monitor is also implemented to watch over currently running APPs. If an APP starts or exits, it will also trigger the scheduler.

% 4 pages
%&latex 
%!TEX root = mobicom2018.tex

\section{Evaluation}
\label{sec.evaluation}

%%%%%%%%%%%%%%%%%%%%%%%%%%%%%%%%%%%%%%%%%%%%%%%%%%%%%%%%%%%%%%%%

%\noindent
%To evaluate {\sysname}, we first present the design choice of six mobile deep learning applications. We then evaluate the performance of {\sysname} in optimizing the each application separately. Lastly, we evaluate the system performance of {\sysname} when six applications are operating concurrently on a mobile device.

%%
%\subsection{Experimental Setup}
%\subsection{Datasets, DNN Models and Mobile Vision Applications}
\subsection{Datasets, DNNs and Applications}
\vspace{0mm}
\subsubsection{Datasets}
%\noindent
%
$\\$ 
To evaluate the generalization capability of {\sysname} on different vision tasks, we select two types of tasks that are among the most important tasks for mobile vision systems.
%
%1) \textit{generic-category object recognition}, and 2) \textit{class-specific object recognition}, .
%are among the need to are representative of common mobile deep learning tasks
%Recognizing various objects in an image is arguably most important task in mobile deep learning applications. 

\vspace{1mm}
\noindent
\textbf{Generic-Category Object Recognition.} 
This type of vision tasks aims to recognize the generic category of an object (e.g., a road sign, a person, or an indoor place). 
%For example, an application is designed to recognize all common objects (e.g., dog or road sign) in daily life. 
%
Without loss of generality, we select 3 commonly used computer vision datasets, each containing a small, a medium, and a large number of object categories respectively, representing an easy, a moderate, and a difficult vision task correspondingly.
\squishlist{
\item{
\textbf{CIFAR-10}~\cite{krizhevsky2009learning}. 
This dataset contains 50K training images and 10K testing images belonging to 10 generic categories of objects. 
} 
\item{
\textbf{ImageNet-50}~\cite{russakovsky2015imagenet}.
This dataset is a subset of the ILSVRC ImageNet.
It contains 63K training images and 2K testing images belonging to top 50 most popular object categories based on the popularity ranking provided by the official ImageNet website. 
%
%For short, we use ImageNet-50 to describe this dataset in the future.
%contains over one million of images of 1,000 classes. 
%Since {\sysname} focuses on deep learning in a mobile environment, it may not be possible for a mobile device (e.g, a smartphone) to encounter all 1,000 classes within a short period of time. Therefore, we select the most popular 50 out of 1,000 classes according to the official popularity percentile provided by ImageNet website. In total, it has over 63K and 2K images in training and validation set. For short, we use ImageNet-50 to describe this dataset in the future.
}
\item{
\textbf{ImageNet-100}~\cite{russakovsky2015imagenet}.
Similar to ImageNet-50, this dataset is a subset of the ILSVRC ImageNet.
It contains 121K training images and 5K testing images belonging to top 100 most popular object categories based on the popularity ranking provided by the official ImageNet website.  
%we select the most popular 100 classes from ImageNet. It contains over 121K and 5K in training and validation set, respectively. 
%For short, we use ImageNet-100 in the future.
}
}\squishend

\vspace{1mm}
\noindent
\textbf{Class-Specific Object Recognition.}
This type of vision tasks aims to recognize the specific class of an object within a generic category (e.g., a stop sign, a female person, or a kitchen). 
%For example, an object belongs to a generic category (e.g., road sign) and specific category recognition is triggered to provide a fine-grained classification among all possible subcategories (e.g., speed limit sign or stop sign). 
%
Without loss of generality, we select 3 object categories: 1) road signs, 2) people, and 3) places, which are commonly seen in mobile settings.
\squishlist{
\item{
\textbf{GTSRB}~\cite{stallkamp2012man}.
This dataset contains over 50K images belonging to 43 classes of road signs such as speed limit signs and stop signs. 
%We choose it as it is representative of automatic driving system applications.
} 
\item{
\textbf{Adience-Gender}~\cite{levi2015age}.
This dataset contains over 14K images of human faces of two genders. 
}
\item{
\textbf{Places-32}~\cite{zhou2017places}.
This dataset is a subset of the Places365-Standard dataset. 
Places365-Standard contains 1.8 million images of 365 scene classes belonging to 16 higher-level categories. 
We select two representative scene classes (e.g., parking lot and kitchen) from each of the 16 higher-level categories and obtain a 32-class dataset that includes over 158K images.
%It contains the top 50 most popular object categories based on the popularity ranking provided by the official ImageNet website. 
%
%For mobile application, we select the most representative 2 scene categories (e.g., parking lot and kitchen) from each of the 16 classes and obtain a 32-class dataset that consist of over 158K images. 
%For short, we use Places-32 in the future.
}
}\squishend

%\item{
% GTSRB~\cite{stallkamp2012man}: It has 60,000 images consisting of animals and vehicles. 
%} 

%%%%%%%%%%%%%%%%%%%%%%%%%%%%%%%%%%%%%%%%%%%%%%%%%%%%%%%%%%%%%%%%

\vspace{-2mm}
\subsubsection{DNN Models}
$\\$ 
To evaluate the generalization capability of {\sysname} on different DNN models, we select two representative DNN models: 1) VGG-16 and 2) ResNet-50. 
VGG-16~\cite{simonyan2014very} is considered as one of the most straightforward DNN models to implement, and thus gains considerable popularity in both academia and industry.
ResNet-50~\cite{he2016deep} is considered as one of the top-performing DNN models in computer vision due to its superior recognition accuracy. 
%In this work, we use VGG-16 and ResNet-50.% which contains 16 and 50 convolutional layers, respectively.

%%%%%%%%%%%%%%%%%%%%%%%%%%%%%%%%%%%%%%%%%%%%%%%%%%%%%%%%%%%%%%%%

\vspace{-2mm}
\subsubsection{Mobile Vision Applications}
$\\$ 
Without loss of generality, we randomly assign CIFAR-10, GTSRB and Adience-Gender to VGG-16; and assign ImageNet-50, ImageNet-100 and Places-32 to ResNet-50 to create six mobile vision applications labeled as \textsf{VC} (i.e., VGG-16 trained on the CIFAR-10 dataset), \textsf{RI-50}, \textsf{RI-100}, \textsf{VS}, \textsf{VG}, and \textsf{RP}, respectively. 
We train and test all the vanilla DNN models and all the descendant models generated by {\sysname} by strictly following the protocol provided by each of the six datasets described above. 

The datasets, DNN models, and mobile vision applications are summarized in Table~\ref{app_sum}.

\begin{table}[t]
\centering
\scalebox{0.74}{
\begin{tabular}{|c|c|c|c|}
\hline
\textbf{Type}                                                                         & \textbf{Dataset}        & \textbf{DNN Model} & \textbf{Mobile Vision Application} \\ \hline
\multirow{3}{*}{\begin{tabular}[c]{@{}c@{}}Generic\\ Category\end{tabular}}  & CIFAR-10       & VGG-16        & \textsf{VC}          \\ \cline{2-4} 
                                                                             & ImageNet-50    & ResNet-50     & \textsf{RI-50}        \\ \cline{2-4} 
                                                                             & ImageNet-100   & ResNet-50     & \textsf{RI-100}      \\ \hline
\multirow{3}{*}{\begin{tabular}[c]{@{}c@{}} Class \\ Specific\end{tabular}} & GTSRB          & VGG-16        & \textsf{VS}           \\ \cline{2-4} 
                                                                             & Adience-Gender & VGG-16        & \textsf{VG}       \\ \cline{2-4} 
                                                                             & Places-32      & ResNet-50     & \textsf{RP}      \\ \hline
\end{tabular}
}
\vspace{1mm}
\caption{Summary of datasets, DNN models, and mobile vision applications used in this work.}
%For example, the mobile vision application \textsf{VC} is built on the VGG-16 model trained on the CIFAR-10 dataset.
\label{app_sum}
\vspace{-10mm}
\end{table}

%%%%%%%%%%%%%%%%%%%%%%%%%%%%%%%%%%%%%%%%%%%%%%%%%%%%%%%%%%%%%%%%

\begin{figure*}[t]
\centering
\includegraphics[scale=0.61]{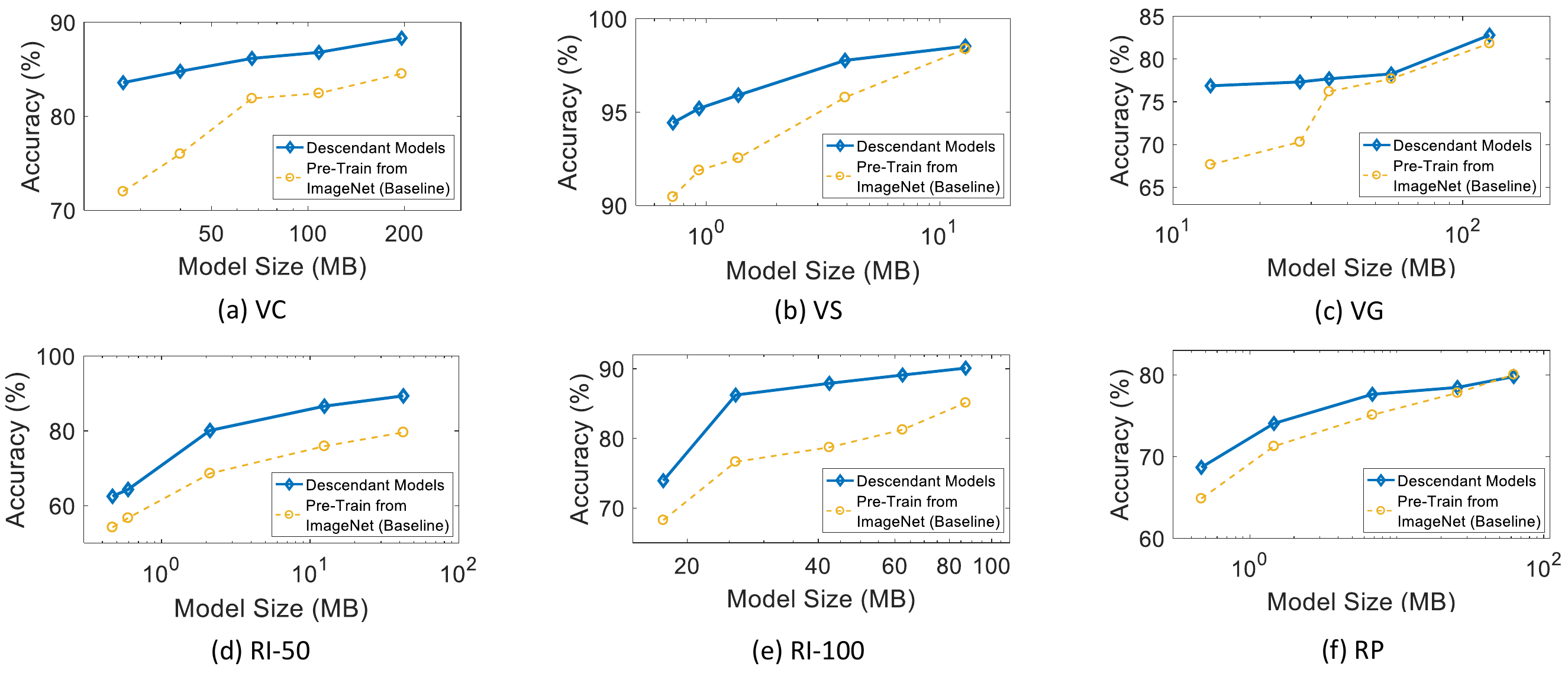}
\vspace{-2mm}
\caption{Top-1 accuracy vs. model size comparison between descendent models and baseline models.
%For better illustration purpose, the horizontal axis is plotted using the logarithmic scale.
%The redundancy zone measures the redundant model size a vanilla model can reduce without accuracy degradation. 
}
\vspace{-1mm}
\label{dia.recovery}
\end{figure*} 

%%%%%%%%%%%%%%%%%%%%%%%%%%%%%%%%%%%%%%%%%%%%%%%%%%%%%%%%%%%%%%%%

\subsection{Performance of Multi-Capacity Model}

In this section, we evaluate the performance of multi-capacity model to demonstrate its superiority listed in \S 3.2.3.

%{\sysname} supports generating nested DNNs with various numbers of included descendant models. 
%We first evaluate the performance of the multi-capacity model in terms of inference accuracy.
%we evaluate model compression and model recovery and quantify its advantage in optimizing resource-accuracy profile at each capacity listed in \S 4.2.3.
%We then quantify its advantages in reducing computational cost, memory footprint, and model switching overhead.

%%%%%%%%%%%%%%%%%%%%%%%%%%%%%%%%%%%%%%%%%%%%%%%%%%%%%%%%%%%%%%%%
\vspace{-2mm}
\subsubsection{Experimental Setup}
$\\$ 
\noindent
\textbf{Selection of Descendant Models.}
Without loss of generality, for each mobile vision application, we generate a multi-capacity model that contains five descendant models.
%We choose these because XXX.
%
These descendant models are designed to have diverse resource-accuracy trade-offs.
We select these descendant models with the purpose to demonstrate that our multi-capacity model enables applications to run even when available resources are very limited.
% during interim shortage of resources.
% even by devices with very limited resource budget.
%This superiority is especially attractive when it comes to devices with limited resource budget.
It should be noted that a multi-capacity model is neither limited to one particular set of resource-accuracy trade-offs nor limited to one particular number of descendant models.
{\sysname} provides the flexibility to design a multi-capacity model based on users' preferences. 

%%%%%%%%%%%%%%%%%%%%%%%%%%%%%%%%%%%%%%%%%%%%%%%%%%%%%%%%%%%%%%%%

\vspace{1mm}
\noindent
\textbf{Baseline.}
%%
% P2: Talk about baseline, and why we choose it as our baseline to compare with.
%To demonstrate their superiority, 
%We compare the generated descendant models with a series of baselines. 
%
To make a fair comparison, we use the same architecture of descendant models for baseline models such that their model sizes and computational costs are identical.
In addition, we pre-trained baseline models on ImageNet dataset and then fine-tuned them on each of the six datasets. 
Pre-training is an effective way to boost accuracy, and thus is adopted as a routine in machine learning community~\cite{oquab2014learning, huh2016makes}.
%achieve state-of-the-art inference accuracy
%reportedly achieves state-of-the-art performance~\cite{oquab2014learning, huh2016makes}.
%
We also trained baseline models without pre-training, and observed that baseline models with pre-training consistently outperform those without pre-training.
Therefore, we only report accuracies of baseline models with pre-training.

%%%%%%%%%%%%%%%%%%%%%%%%%%%%%%%%%%%%%%%%%%%%%%%%%%%%%%%%%%%%%%%%

\begin{figure*}[t]
\centering
\includegraphics[scale=0.365]{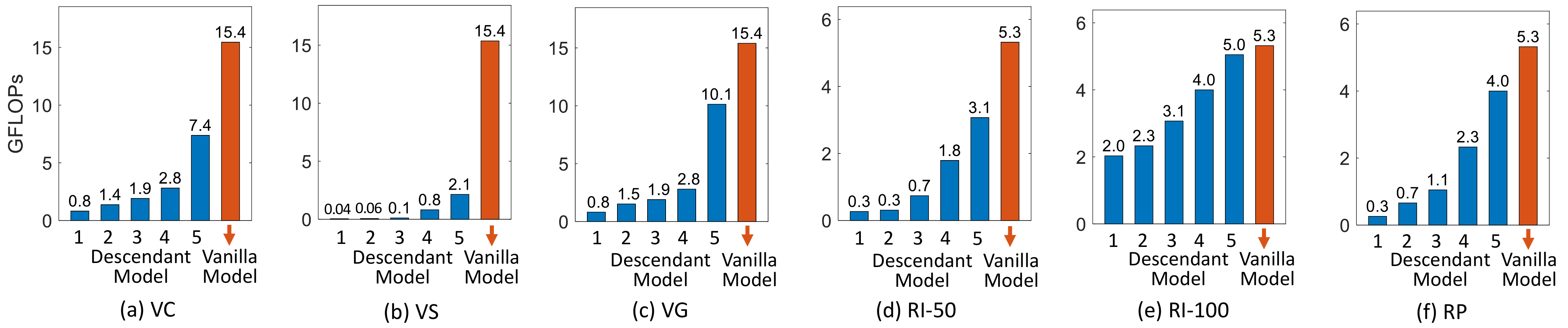}
\vspace{-2.5mm}
\caption{Computational cost comparison between descendant models and vanilla models. 
%FPS results are obtained by running applications on a Samsung Galaxy S8 smartphone.
}
\vspace{-3mm}
\label{dia.profile}
\end{figure*} 

%%%%%%%%%%%%%%%%%%%%%%%%%%%%%%%%%%%%%%%%%%%%%%%%%%%%%%%%%%%%%%%%

\vspace{-2mm}
\subsubsection{Optimized Resource-Accuracy Trade-offs}
$\\$ 
%
% P1: Talk about descendant models: why we generate those 5 ones.
%One key advantage of multi-capacity model is the generation of descendant models with optimized resource-accuracy trade-offs.
% P3: Describe the figure.
Figure~\ref{dia.recovery} illustrates the comparison between descendent models and baseline models across six mobile vision applications. 
For each application, we show the top-1 accuracies of both descendant models and baseline models as a function of model size. 
%Each diamond marker on the blue curve corresponds to 
%The orange triangle symbols represent vanilla models. 
%The redundancy zone measures the redundant model size a vanilla model can reduce without accuracy degradation.
%
For better illustration purpose, the horizontal axis is plotted using the logarithmic scale.
%For example, Figure~\ref{dia.recovery}(a) shows the comparison of \textsf{VC}.
%As shown, the first descendant model (the leftmost blue diamond) is 83.58\% in accuracy and 26.5 MB in model size.
%Meanwhile, the corresponding baseline model (the leftmost yellow circle) is 71.91\% in accuracy, with exactly the same model size (26.5 MB).
%We highlight the model size difference between vanilla model (red triangle) and the largest descendant model (marked as ``Redundancy Zone'').

%%
% P4 and P5: Describe our two findings and their indications.
%
We have two key observations from the result. 
%%
%First, we observe that descendant models are able to recover to the accuracy of DNN Pruning models. 
%%
%%$0.54 \pm 2.43 \%$
%%$0.73 \pm 2.76 \%$
%On average, the accuracy of descendant models is 0.54\% higher than DNN Pruning models across six applications. 
%%
%Interestingly, in some applications, descendant models outperform DNN Pruning.
%%
%For example, in \textsf{RI-100}, the descendant models are able to outperform DNN Pruning by an average 5.41\%.
%%It indicates that descendant models are able to recover to DNN Pruning models.
%%As such, the fact the DNN Pruning models have state-of-the-art performance demonstrated in \S 4.1.4 makes descendant models extremely reliable.
%%
%One reason is that descendant models follow the filter pruning roadmap provided by DNN Pruning and hence are able to add the most important filters.
%%
First, we observe that descendant models consistently achieve higher accuracies than baseline models at every model size across all the six applications.
On average, descendant models achieve 4.98\% higher accuracy than baseline models. 
% our TRR approach effectively identifies and prunes less important filters and thus the 
This indicates that our descendant model at each capacity is able to deliver state-of-the-art inference accuracy under a given memory budget.
Second, we observe that smaller descendant models outperform baseline models more than larger descendant models.
On average, the two smallest descendant models achieve 6.68\% higher accuracy while the two largest descendant models achieve 3.72\% higher accuracy compared to their corresponding baseline models.
This is because our TRR approach is able to preserve important filters while pruning less important ones.
Despite having a small capacity, a small descendant model benefits from these important filters while the corresponding baseline model does not.
Figure~\ref{dia.profile} shows the computational costs of five descendant models and the corresponding vanilla models of the six applications in GFLOPs (i.e., GigaFLOPs).
% Point#1: 
As shown, all descendant models have less GFLOPs than the corresponding vanilla models. 
%
%This result demonstrates the benefits of filter pruning on reducing the computational intensity of deep learning models. 
%In particular, the smallest descendant model have $\times 18.8$ (\textsf{VC}), $\times 384.0$ (\textsf{VS}), $\times 19.0$ (\textsf{VG}), $\times 19.7$ (\textsf{RI-50}), $\times 2.6$ (\textsf{RI-100}), and $\times 20.5$ (\textsf{RP}) less FLOP.
%Compared to vanilla model, the descendant models have up to $\times 18.8$, $\times 384.0$, $\times 19.0$, $\times 19.7$, $\times 2.6$, and $\times 20.5$ speedup of application VC, VS, VG, RI-50, RI-100 and RP, respectively. 
%
This result indicates that our filter pruning approach is able to effectively reduce the computational costs across six applications, demonstrating the generalization of our filter pruning approach on different deep learning models trained on different datasets.

%%
%As another pruning approach, pruning model parameters could also reduce model size~\cite{han2015learning}. 
%However, this method does not reduce FLOP effectively~\cite{luo2017thinet, molchanov2016pruning, li2016pruning}, making it less attractive in generating descendant models with diverse resource-accuracy profiles.
%In contrast, our filter pruning approach is able to reduce model size and FLOP simultaneously.

%This indicates {\sysname} is able to generate models with a wide range of FLOP, meeting the requirements of different resource budgets (i.e., computational budget) in a mobile environment.
%
% Point#2: within each app, we achieve higher FPS. 
% NO Point#3: we are better than GPU, and we are confident we can achieve better when deploying in GPU. 
%COMPARE TO state-of-the-art results from DeepMon which runs on mobile GPU.
%Emphasize we can achieve similar and better results using mobile CPU.

%%%%%%%%%%%%%%%%%%%%%%%%%%%%%%%%%%%%%%%%%%%%%%%%%%%%%%%%%%%%%%%%

%\subsection{Performance of Multi-Capacity Model}

%%
%In this section, we evaluate the performance of multi-capacity model and quantify its advantages in reducing memory footprint and model switching overhead listed in \S 4.2.3.

%\noindent
%During recovery stage, {\sysname} adopts freezing techniques to allow weight sharing among descendants. 
%As a result, a fully recovered model encapsulates all the previously recovered models without increasing its size. 
%This leads to two benefits: 1) compression of model size; and 2) overhead reduction during model switching.

\subsubsection{Reduction on Memory Footprint} 
$\\$ 
Another key feature of multi-capacity model is sharing parameters among its descendant models.
To quantify the benefit of parameter sharing on reducing memory footprint, we compare the model size of multi-capacity model with the accumulated model size of the five descendant models as if they were independent. 
This mimics traditional model variants that are used in existing mobile deep learning systems. 

Table~\ref{tab.compress} lists the comparison results across the six mobile vision applications.
Obviously, the model size of the multi-capacity model is smaller than the corresponding accumulated model size for each application.
Moreover, deep learning model with larger model size benefits more from parameter sharing.
%depending on the size of the deep learning model, the reduced memory footprint varies.
%
For example, \textsf{VC} has the largest model size across the six applications. 
With parameter sharing, it achieves a reduced memory footprint of 241.5 MB.
Finally, if we consider running all the six applications concurrently, the multi-capacity model achieves a reduced memory footprint of 587.4 MB, demonstrating the enormous benefit of multi-capacity model on memory footprint reduction.

%the size of the multi-capacity of \textsf{VC} is 196.0 MB and the accumulated model size of five descendant models included in this multi-capacity model is 437.5 MB. 

%Specifically, we compare the size of the multi-capacity model to the accumulated model size of all the descendant models.
%For example, the size of the multi-capacity of \textsf{VC} is 196.0 MB and the accumulated model size of five descendant models included in this multi-capacity model is 437.5 MB.
%As such, the multi-capacity model reduces the memory footprint by 241.5 MB, enabling a $\times 2.2$ compression ratio compared to the five descendant models.
%Overall, the multi-capacity model achieves up to $\times 2.6$ compression ratio compared to the accumulated model size across six applications.
%This is because all the descendant models included in the multi-capacity share parameters among each other due to our freeze-and-grow training paradigm.
%As a result, it effectively eliminates the need of carrying potentially a large number of models with different capacities.

\begin{table}[t]
\centering
\scalebox{0.72}{
\begin{tabular}{|c|c|c|c|}
\hline
\multicolumn{1}{|l|}{\textbf{Application}} & \textbf{\begin{tabular}[c]{@{}c@{}}Multi-Capacity\\  Model Size (MB)\end{tabular}} & \textbf{\begin{tabular}[c]{@{}c@{}}Accumulated\\  Model Size (MB)\end{tabular}} & \textbf{\begin{tabular}[c]{@{}c@{}}Reduced Memory \\  Footprint (MB)\end{tabular}} \\ \hline
\textsf{VC} & 196.0 & 437.5 & 241.5 \\ \hline
\textsf{VS} & 12.9 & 19.8 &  6.9 \\ \hline
\textsf{VG} & 123.8 & 256.0 &  132.2  \\ \hline
\textsf{RI-50} & 42.4 & 58.1 & 15.7 \\  \hline
\textsf{RI-100} & 87.1 & 243.5 &  156.4 \\ \hline
\textsf{RP} & 62.4 & 97.1 &  34.7\\  \hline
\textsf{All Included} & 524.6 & 1112.0  & 587.4  \\ \hline
\end{tabular}
}
\vspace{1mm}
\caption{Benefit of multi-capacity model on memory footprint reduction.}
\label{tab.compress}
\vspace{-10mm}
\end{table}

\begin{figure*}[t]
	\centering
	\includegraphics[scale=0.353]{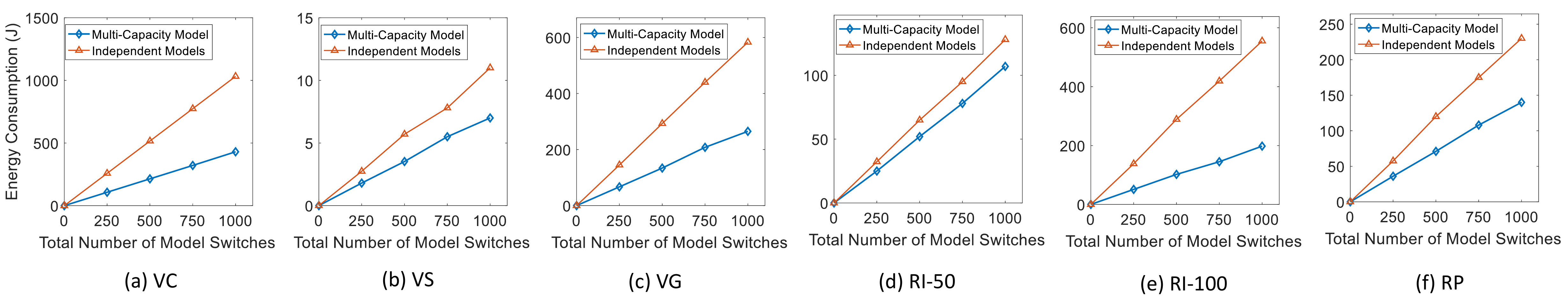}
	\vspace{-3mm}
	\caption{Model switching energy consumption comparison between multi-capacity models and independent models. The energy consumption is measured on a Samsung Galaxy S8 smartphone.}
	\vspace{-2mm}
	\label{dia.switch}
\end{figure*} 
%The energy consumption is measured on a Samsung Galaxy S8 smartphone (unit: J).
%%%%%%%%%%%%%%%%%%%%%%%%%%%%%%%%%%%%%%%%%%%%%%%%%%%%%%%%%%%%%%%%

%\vspace{1mm}
%\noindent
\subsubsection{Reduction on Model Switching Overhead} 
$\\$ 
%
%Because of parameter sharing, during model switching, multi-capacity model only needs to page in and page out a very small portion of the deep learning models.
%
Another benefit of parameter sharing is reducing the overhead of model switching when the set of concurrent applications changes.
To quantify this benefit, we consider all the possible model switching cases among all the five descendant models of each multi-capacity model, and calculate the average page-in and page-out overhead for model upgrade and model downgrade, respectively.
We compare it with the case where descendant models are treated as if they were independent, which again mimics traditional model variants.

Table~\ref{tab.upgrade} lists the comparison results of all six mobile vision applications for model upgrade in terms of memory usage.
%since switching independent models requires to page in and page out the entire models
As expected, the average page-in and page-out memory usage of independent models during model switching is larger than multi-capacity models for every application. 
This is because during model switching, independent models need to page in and page out the \textit{entire} models while multi-capacity models only need to page in a very small portion of the models.
It should be noted that the page-out overhead of multi-capacity model during model upgrade is zero. 
This is because the descendant model with smaller capability is part of the descendant model with larger capability, and thus it does not need to be paged out. 

Table~\ref{tab.downgrade} lists the comparison results of all six applications for model downgrade.
Similar results are observed.
The only difference is that during model downgrade, the page-in overhead of multi-capacity model is zero. 

%%%%%%%%%%%%%%% new addition.
%To further demonstrate the superiority in reducing model switching overhead, we evaluate its impact on energy consumption.
%
Besides memory usage, we also quantify the benefit on reducing the overhead of model switching in terms of energy consumption.
%Again, we compare it with the case where descendant models are treated as if they were independent.
%
Specifically, we measured energy consumed by randomly switching models for 250, 500, 750, and 1,000 times using descendant models and independent models, respectively.
Figure~\ref{dia.switch} shows the comparison results across all six mobile vision applications.
%the comparison between switching multi-capacity models and independent models of six mobile vision applications.
%We show the energy consumption as a function of total number of model switches in the unit of J (i.e., Joule).
%For each application, we repeat the experiment 100 times and report the average result.
%
As expected, energy consumed by switching multi-capacity model is lower than switching independent models for every application.
%Specifically, multi-capacity model reduces energy consumption by as much as $2.8\times$.
%Moreover, we observe that switching between larger models reduces more energy consumption.
%
This benefit becomes more prominent when the model size is large.
For example, the size of the largest descendant model of \textsf{VC} and \textsf{VS} is 196.0 MB and 12.9 MB, respectively. 
The corresponding energy consumption reduction for every 1,000 model switches is 602.1 J and 4.0 J, respectively.
%This indicates that applications with larger models benefit more from multi-capacity model switching.

%%
%Taken together, the generated multi-capacity model is able to significantly reduce model page-in and page-out overhead in terms of both memory usage and energy consumption.
%, making model switching extremely efficient. 
%This also leads to a significant reduction on energy consumption, which is important for mobile systems.
Taken together, the generated multi-capacity model is able to significantly reduce the overhead of model switching in terms of both memory usage and energy consumption.
The benefit becomes more prominent when model switching frequency increases.
This is particularly important for memory and battery constrained mobile systems.

\begin{table}[h]
\centering
\scalebox{0.73}{
\begin{tabular}{|c|c|c|c|c|}
\hline
\multirow{2}{*}{\textbf{Application}} & \multicolumn{2}{c|}{\textbf{\begin{tabular}[c]{@{}c@{}}Multi-Capacity Model \\ Upgrade Overhead (MB)\end{tabular}}} & \multicolumn{2}{c|}{\textbf{\begin{tabular}[c]{@{}c@{}}Independent Models \\ Upgrade Overhead (MB)\end{tabular}}} \\ \cline{2-5} 
 & Page-In & Page-Out & Page-In & Page-Out \\ \hline
\textsf{VC} & 81.4 & 0 & 128.2 & 46.8 \\ \hline
\textsf{VS} & 1.3 & 0 & 1.7 & 0.3 \\ \hline
\textsf{VG} & 50.0 & 0 & 76.2 & 26.2 \\ \hline
\textsf{RI-50} & 19.2 & 0 & 21.2 & 2.0 \\ \hline
\textsf{RI-100} & 38.3 & 0 & 67.9 & 29.5 \\ \hline
\textsf{RP} & 26.4 & 0 & 34.9 & 4.6 \\ \hline
\end{tabular}
}
\vspace{1mm}
\caption{Benefit of multi-capacity model on model switching (model upgrade) in terms of memory usage.}
\label{tab.upgrade}
\vspace{-6mm}
\end{table} 

\begin{table}[h]
\centering
\scalebox{0.71}{
\begin{tabular}{|c|c|c|c|c|}
\hline
\multirow{2}{*}{\textbf{Application}} & \multicolumn{2}{c|}{\textbf{\begin{tabular}[c]{@{}c@{}}Multi-Capacity Model \\ Downgrade Overhead (MB)\end{tabular}}} & \multicolumn{2}{c|}{\textbf{\begin{tabular}[c]{@{}c@{}}Independent Models \\ Downgrade Overhead (MB)\end{tabular}}} \\ \cline{2-5} 
 & Page-In & Page-Out & Page-In & Page-Out \\ \hline
\textsf{VC} & 0 & 81.4 & 46.8 & 128.2 \\ \hline
\textsf{VS} & 0 & 1.3 & 0.3 & 1.7 \\ \hline
\textsf{VG} & 0 & 50.0 & 26.2 & 76.2 \\ \hline
\textsf{RI-50} & 0 & 19.2 & 2.0 & 21.2 \\ \hline
\textsf{RI-100} & 0 & 38.3 & 29.5 & 67.9 \\ \hline
\textsf{RP} & 0 & 26.4 & 4.6 & 34.9 \\ \hline
\end{tabular}
}
\vspace{1mm}
\caption{Benefit of multi-capacity model on model switching (model downgrade) in terms of memory usage.}
\label{tab.downgrade}
\vspace{-6mm}
\end{table}

\subsection{Performance of Resource-Aware Scheduler}
%\subsection{Runtime Performance of Concurrent Applications}
%In this part, we first evaluate the runtime performance of the six mobile applications implemented in a mobile device under two state-of-the-art scheduling schemes. 
%We also evaluate the impact of {\sysname} on energy consumption caused by model switching.
%%%%%%%%%%%%%%%%%%%%%%%%%%%%%%%%%%%%%%%%%%%%%%%%%%%%%%%%%%%%%%%%

\subsubsection{Experimental Setup}
%%%%%%%%%%%%%%%%%%%%%%%%%%%%%%%%%%%%%%%%%%%%%%%%%%%%%%%%%%%%%%%%
%
$\\$ 
\noindent
\textbf{Deployment Platforms.}
%
%We have implemented those six representative applications using {\sysname} on Samsung Galaxy S8 which is running the latest Android OS 7.0.
We implemented {\sysname} and the six mobile vision applications on three smartphones: Samsung Galaxy S8, Samsung Galaxy S7, and LG Nexus 5, all running Android OS 7.0.
We used Monsoon power monitor~\cite{monsoon} to measure the power consumption.
%While measuring the power consumption, we terminated all irrelevant services of the Android OS and turned off the smartphone screen.
%%
We have achieved consistent results across all three smartphones.
%Due to the limitation of space, 
%
Here we only report the best results obtained from Samsung Galaxy S8.
%Since we achieved the best results from Galaxy S8, here we only report the results obtained from Galaxy S8.

%
%Specifically, it is powered by an octa-core CPU with up to 2.45 GHz frequency, a GPU with up to 710 MHz frequency and 4 GB LPDDR4 RAM.
%Specifically, Samsung Galaxy S8 is powered by an octa-core CPU with up to 2.45 GHz frequency and 4 GB LPDDR4 RAM.
%We used the Monsoon power monitor~\cite{monsoon} to measure the power consumption of Galaxy S8.
%It is worthwhile to note that we have also implemented {\sysname} on Samsung Galaxy S7 and LG Nexus 5 smartphones, and observed similar results.
%We only report results on Galaxy S8 as it is representative of the other two smartphones.
%
%.
%{\xiao{should we mention the how much cache we use in the scheduler? }}
%%%%%%%%%%%%%%%%%%%%%%%%%%%%%%%%%%%%%%%%%%%%%%%%%%%%%%%%%%%%%%%%

%%%%%%%%%%%%%%%%%%%%%%%%%%%%%%%%%%%%%%%%%%%%%%%%%%%%%%%%%%%%%%%%

\vspace{1mm}
\noindent
\textbf{Baseline.}
We used the status quo approach (i.e., resource-agnostic) which uses fixed resource-accuracy trade-off as the baseline.
It uses the model located at the ``knee'' of every yellow curve in Figure \ref{dia.recovery}.
This is the one that achieves the best resource-accuracy trade-off among all the model variants.
%We use the model located at the ``knee'' of every yellow curve in Figure \ref{dia.recovery} as our baseline for each mobile vision application.
%This is the model that offers the \textit{best} resource-accuracy trade-off among all the model variants.
%

%We also implemented the same six mobile vision applications where only \textit{one} independent baseline model is selected for each of the six mobile vision applications. 
%We use this approach as our baseline as it is adopted by state-of-the-art mobile deep learning approaches.
%To make a fair comparison, we select independent baseline models from the knees of the yellow curves in Figure \ref{dia.recovery} to get best resource-accuracy trade-off, following the practice of~\cite{mcdnn}.

%We use traditional model deep learning framework as our baseline, where only one independent model is used for each application.
%%
%The independent model is pre-trained and ImageNet and fine-tuned on the corresponding dataset, the same as the baseline in Section 5.2.
%%
%Following the same way in \cite{mcdnn}, we pick models from the knees of the yellow curves in Figure \ref{dia.recovery} for each application.
%
%In particular, the independent models are the baseline models introduced \S 5.2.1.
%

%%%%%%%%%%%%%%%%%%%%%%%%%%%%%%%%%%%%%%%%%%%%%%%%%%%%%%%%%%%%%%%%

\vspace{1mm}
\noindent
\textbf{Benchmark Design.}
To compare the performance between our resource-aware approach and the resource-agnostic status quo approach, we have designed a benchmark that emulates runtime application queries in diverse scenarios. 
%
%In this benchmark, an application will be generated or killed with a certain probability at every time instant randomly. At most six applications are allowed to exist at a time.
%
Specifically, our benchmark creates a new application or kills a running application with certain probabilities at every second.
The number of concurrently running applications is from 2 to 6.
%It allows two to six applications running concurrently to simulate the variation of application query.
%
%It starts with an arbitrary number (e.g., three) of applications running concurrently.
%
The maximum available memory to run concurrent applications is set to $400$ MB.
Each simulation generated by our benchmark lasts for $60$ seconds.
We repeat the simulation 100 times and report the average runtime performance.

\begin{figure}[t]
\centering
\includegraphics[scale=0.45]{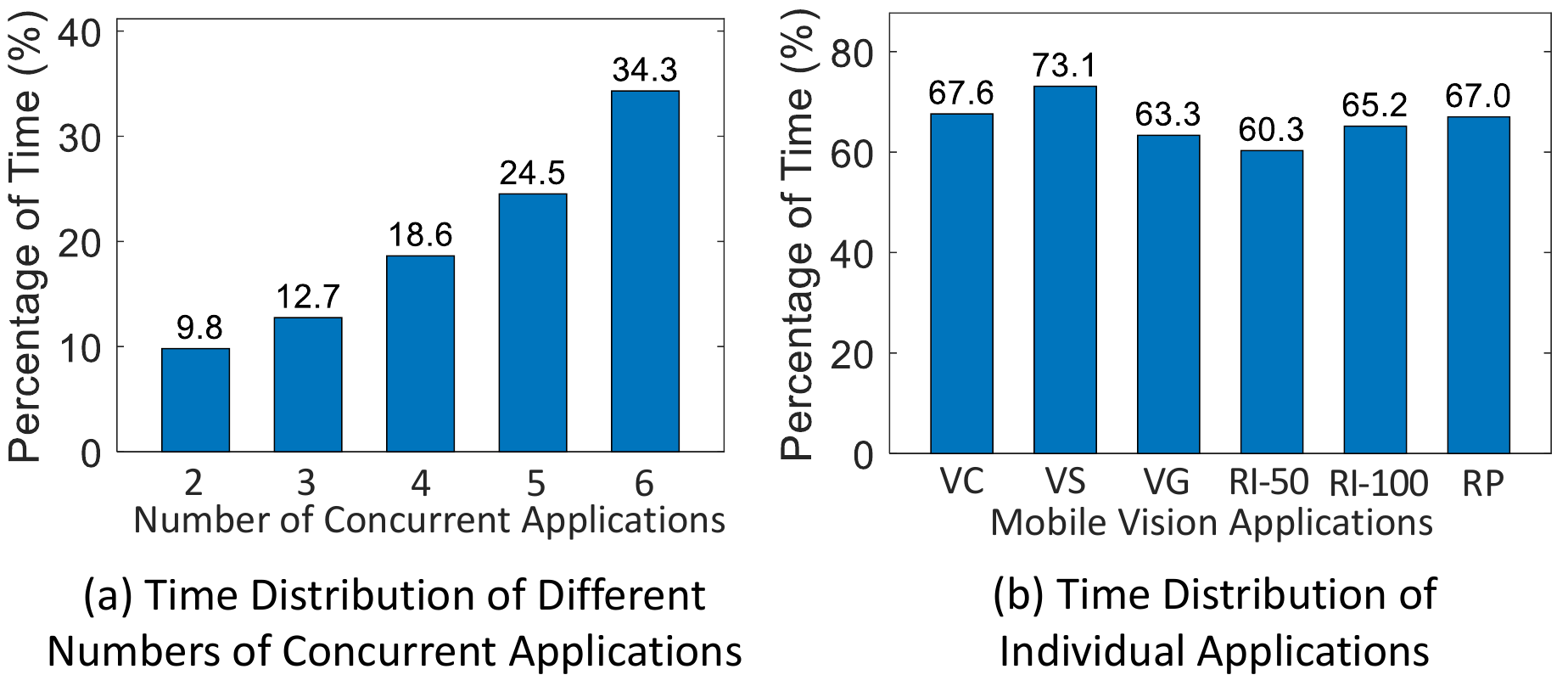}
\vspace{-6mm}
\caption{Benchmark profile.}
\vspace{-5mm}
\label{dia.benchmark}
\end{figure}

Figure~\ref{dia.benchmark} shows the profile of all accumulated simulation traces generated by our benchmark.
Figure~\ref{dia.benchmark}(a) shows the time distribution of different numbers of concurrent applications.
As shown, the percentage of time when two, three, four, five and six applications running concurrently is 9.8\%, 12.7\%, 18.6\%, 24.5\% and 34.3\%, respectively. 
%It shows that our benchmark has covered all the numbers of concurrent applications.
%
Figure~\ref{dia.benchmark}(b) shows the running time distribution of each individual application.
As shown, the running time of each application is approximately evenly distributed, indicating our benchmark ensures a reasonably fair time share among all six applications.

%
%In this way, the system can be comprehensively evaluated in every concurrent scenario for each of the six applications. 

%
% XX\%, XX\%, XX\%, XX\%, XX\% of the time the system needs to run 2,3,4,5,6 applications respectively.
%Most of the time system needs to run 6 concurrent applications because we want to evaluate the system under stress load.
%%
%Figure XXX shows the average running time for each application. 
%We can see that the simulation equally covers inferences and the switches among the six applications.
%%

%

%
%
%To simulate runtime application query, we randomly 
%
%
%To simulate internal variation, we randomly generate a APP from six applications or destroy one of the running APPs. The number of running APPs are constrained to range from two and six.
%%
%To simulate the external variation, we impose an random variation on available computing resource $\delta cpu = \gamma CPU$ where $\gamma ~ uniform(0,0.1)$.
%%
%%
%Meanwhile, we set the maximum memory available for concurrent applications at 400 MB.
%%
%We also implement the MinTotalCost and MinMaxCost scheduler using cached greedy heuristic method and adopt an aggressive caching strategy by caching the previous $90\%$ scheme.
%%
%Each simulation lasted for 60 seconds. We run 100 simulations and reported the average result.

%%%%%%%%%%%%%%%%%%%%%%%%%%%%%%%%%%%%%%%%%%%%%%%%%%%%%%%%%%%%%%%%

\vspace{1mm}
\noindent
\textbf{Evaluation Metrics.} 
We use the following two metrics to evaluate the scheduling performance.
%The following metrics are used to measure the overall quality and speed of six concurrent mobile vision applications. 
\vspace{-1mm}
\squishlist{
\item{\textbf{Inference Accuracy Gain.} 
Since the absolute top-1 accuracies obtained by the six applications are not in the same range, we thus use accuracy gain over the baseline within each application as a more meaningful metric.
%In other words, accuracy gain for each application is obtained by subtracting the accuracy of corresponding smallest baseline model.
%For example, the accuracy of the smallest descendant model of \textsf{VC} is $83.5\%$ and accuracy of the smallest baseline model of \textsf{VC} is $72.0\%$.
%The accuracy gain of this descendant model can be calculated as $83.5\% - 72.0\% = 11.5\%$.
%When the applications are running concurrently, we report average accuracy gain as a measurement of overall inference quality. 
}
\vspace{-1mm}
\item{
\textbf{Frame Rate Speedup.}
Similarly, since the absolute frame rate achieved depends on the number of concurrently running applications, we thus use frame rate speedup over the baseline as a more meaningful metric.
% to measure the real-time performance. 
%It is a standard metric for mobile vision systems.
%Frame rate is inversely proportional to inference latency.
%The lower the inference latency is, the higher the frame rate is.
}
}\squishend
%
%These two metrics altogether reflect the overall inference quality and speed of a vision application.

%%%%%%%%%%%%%%%%%%%%%%%%%%%%%%%%%%%%%%%%%%%%%%%%%%%%%%%%%%%%%%%%

%\subsubsection{Performance under State-of-the-Art Scheduling Schemes}
%\noindent
%In this part, we first compare the accuracy and FPS trade-off performance between {\sysname} and baseline. We then compare the energy consumption under the same average accuracy gain.

%%%%%%%%%%%%%%%%%%%%%%%%%%%%%%%%%%%%%%%%%%%%%%%%%%%%%%%%%%%%%%%%

%%%%%%%%%%%%%%%%%%%%%%%%%%%%%%%%%%%%%%%%%%%%%%%%%%%%%%%%%%%%%%%%

\vspace{-2mm}
\subsubsection{Improvement on Inference Accuracy and Frame Rate}
\begin{figure}[t]
\includegraphics[scale=0.58]{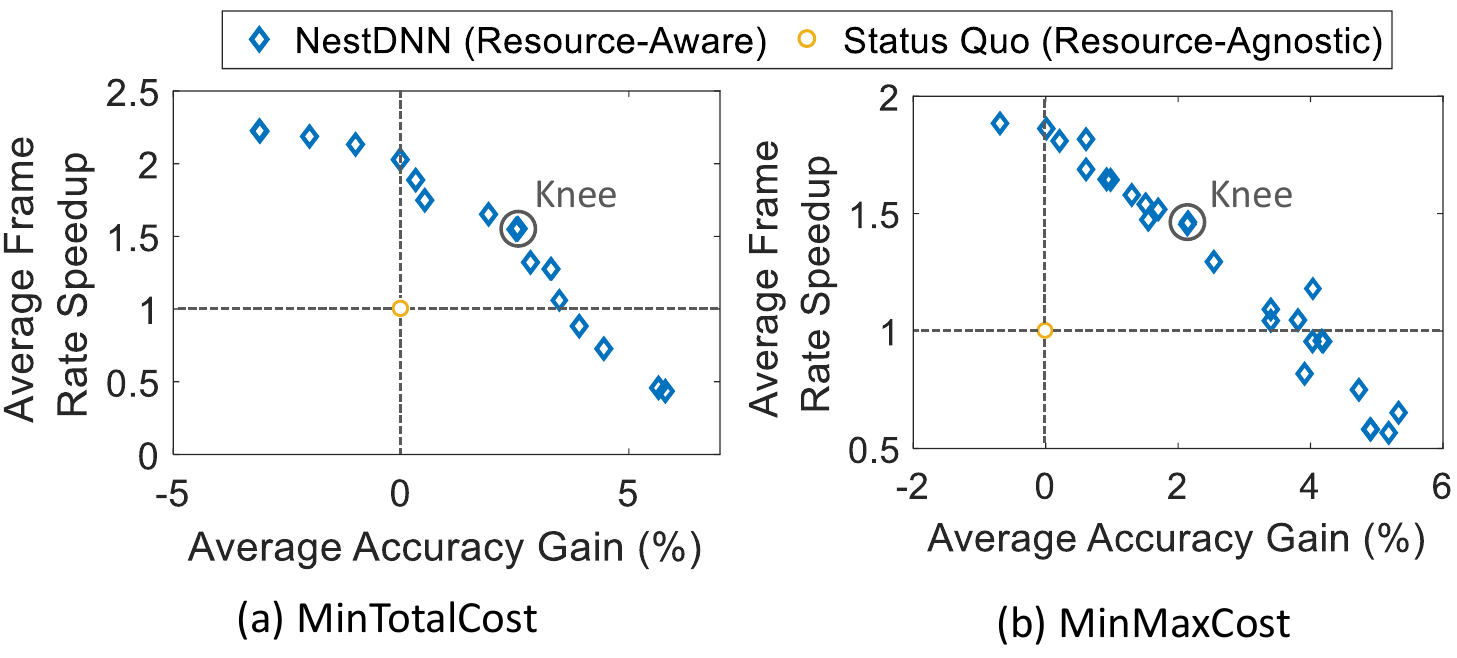}
\centering
\vspace{-7mm}
\caption{Runtime performance comparison between {\sysname} (resource-aware) and status quo (resource-agnostic) under (a) MinTotalCost and (b) MinMaxCost.}
\vspace{-1mm}
\label{dia.scheduler-acc}
\end{figure}

%$\\$ 
%
%%%%give an impression first
%Compared to baseline, {\sysname} is able to provide the same quality inferences with around $2.0\times$ FPS.
%
%%%% desribe them in detail
%
Figure~\ref{dia.scheduler-acc}(a) compares the runtime performance between {\sysname} and the baseline under the MinTotalCost scheduling scheme.
The yellow circle represents the runtime performance of the baseline.
Each blue diamond marker represents the runtime performance obtained by scheduling with a particular $\alpha$ in the cost function defined in Equation (2).

We have two key observations from the result.
First, by adjusting the value of $\alpha$, {\sysname} is able to provide various trade-offs between inference accuracy and frame rate, which the status quo approach could not provide.
%, while baseline can only provide a fix accuracy-FPS output.
%As shown, {\sysname} trade-off between average accuracy gain and average FPS freely .
%In contrast, the baseline is fixed and offers no trade-off between accuracy and frame rate.
% because only one model is used in each application and thus it cannot trade-off accuracy-FPS by switching model.
%  
%This is because multi-capacity models of {\sysname} consist of five descendant models with a unique resource-accuracy trade-off. 
%
Second, the vertical and horizontal dotted lines altogether partition the figure into four quadrants.
The upper right quadrant represents a region where {\sysname} achieves both higher top-1 accuracy and frame rate than the status quo approach.
%that has both higher accuracy gain and higher frame rate compared to the baseline.
%As shown, there are many blue diamond markers locating at this upper right quadrant.
%At each of those blue diamond markers, {\sysname} is able to achieve both higher accuracy gain and higher frame rate compared to the baseline.
%
In particular, we select 3 blue diamond markers within the upper right quadrant to demonstrate the runtime performance improvement achieved by {\sysname}.
Specifically, when {\sysname} has the same average top-1 accuracy as the baseline, {\sysname} achieves $2.0 \times$ average frame rate speedup compared to the baseline. 
When {\sysname} has the same average frame rate as the baseline, {\sysname} achieves $4.1 \%$ average accuracy gain compared to the baseline. 
Finally, we select the ``knee'' of the blue diamond curve, which offers the best accuracy-frame rate trade-off among all the $\alpha$.
At the ``knee'', {\sysname} achieves $1.5 \times$ average frame rate speedup and $2.6 \%$ average accuracy gain compared to the baseline.

%we observe that {\sysname} outperforms the baseline in terms of both accuracy and FPS.
%Concurrent applications that achieve high average accuracy gain at a high average FPS (top right corner; marked as 'best') are considered ideal.
%As shown, several blue diamonds ({\sysname}) are located at the upper right of the yellow circle (baseline). 
%This indicates that {\sysname} is able to achieve higher accuracy and higher FPS than baseline.
%This is because each descendant model generated by {\sysname} is able to deliver the state-of-the-art accuracy under a given resource budget.

%%
Figure~\ref{dia.scheduler-acc}(b) compares the runtime performance between {\sysname} and the baseline under the MinMaxCost scheduling scheme.
%As illustrated, similar results are achieved.
%We observe similar results that {\sysname} is able to provide more flexible accuracy-FPS trade-offs and outperforms baseline in terms of accuracy and FPS at the same time.
%The only difference is that {\sysname} under MinMaxCost allows for a narrower range of trade-offs than under MinTotalCost.
%This is because MinMaxCost prefers a balanced resource allocation among applications, and thus the runtime performance is more evenly distributed. 
%
When {\sysname} has the same average top-1 accuracy gain as the baseline, {\sysname} achieves $1.9 \times$ average frame rate speedup compared to the baseline. 
When {\sysname} has the same average frame rate as the baseline, {\sysname} achieves $4.2 \%$ average accuracy gain compared to the baseline. 
At the ``knee'', {\sysname} achieves $1.5 \times$ average frame rate speedup and $2.1 \%$ average accuracy gain compared to the baseline.

%\vspace{3mm}

\begin{figure}[t]
\centering
\includegraphics[scale=0.58]{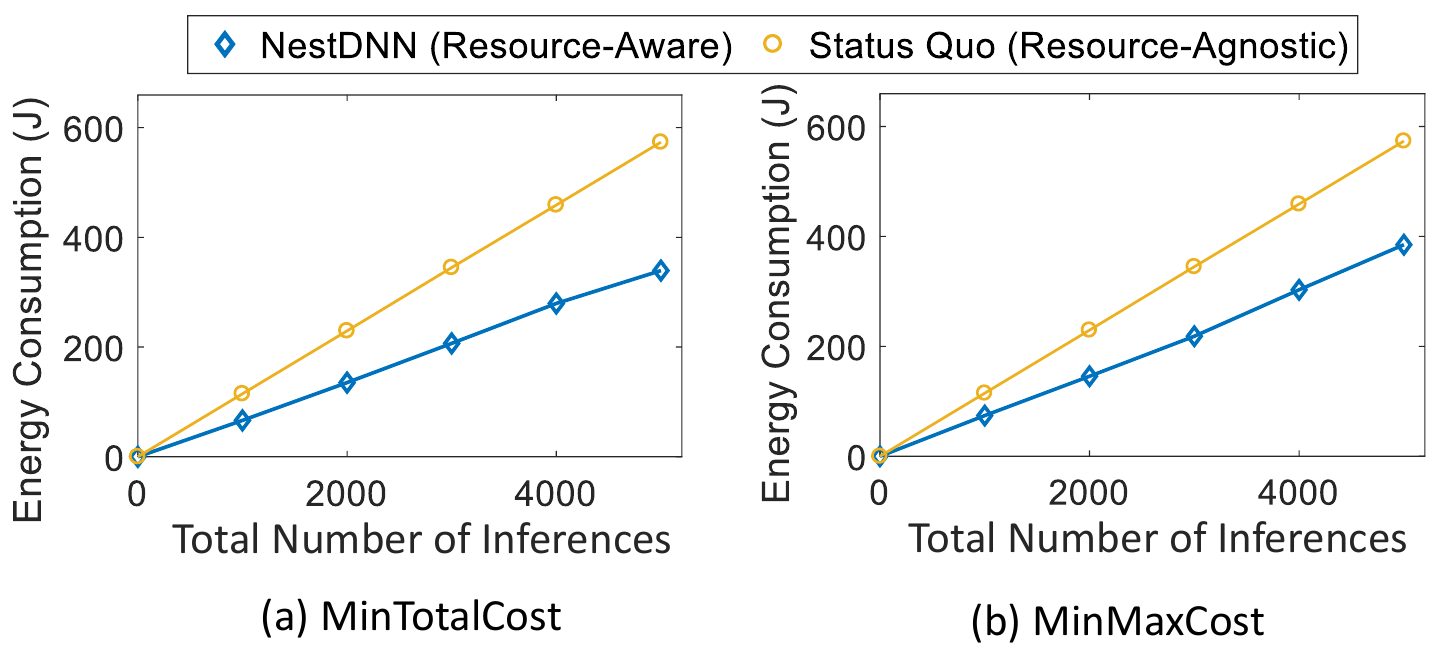}
\vspace{-7mm}
\caption{Energy consumption comparison between {\sysname} (resource-aware) and status quo (resource-agnostic) under (a) MinTotalCost and (b) MinMaxCost.}
\vspace{-3mm}
\label{dia.minmax}
\end{figure} 

\subsubsection{Reduction on Energy Consumption}
$\\$ 
Besides improvement on inference accuracy and frame rate, {\sysname} also consumes less energy.
%Finally, we evaluate the inference energy consumption of {\sysname} at the ``knee'' circled in Figure~\ref{dia.scheduler-acc}.
%To make a fair comparison, we select the case where {\sysname} delivers the \textit{same} average accuracy gain as the baseline.
%Reflected in Figure~\ref{dia.mintotal}(a), the blue diamond symbol linked with baseline by dotted line is selected.
%Specifically, the average FPS of this case is 2.49, which is $2.0 \times$ compared to the baseline.
%Specifically, we evaluate the inference energy consumption of {\sysname} at the ``knee''.
%%
Figure~\ref{dia.minmax}(a) compares the energy consumption between {\sysname} at the ``knee'' and the baseline under the MinTotalCost scheduling scheme.
%Again, the blue diamonds and yellow circles represent the {\sysname} and baseline, respectively.
%
Across different numbers of inferences, {\sysname} achieves an average $1.7 \times$ energy consumption reduction compared to the baseline.
%As shown, under the same number of inferences, {\sysname} consumes lower energy compared to the baseline.
%In particular, with 4000 inferences, compared to the baseline, {\sysname} reduces energy consumption by $ 2.1\times$. 
%%
Similarly, Figure~\ref{dia.minmax}(b) shows the comparison under the MinMaxCost scheduling scheme.
%Similar results are observed.
%Specifically in this comparison, {\sysname} enjoys a $1.9 \times$ speedup compared to baseline.
{\sysname} is able to achieve an average $1.5 \times$ energy consumption reduction compared to the baseline.
%Meanwhile, {\sysname} reduces energy consumption by $1.8 \times$. 

% 1 page
\vspace{-1mm}
%!TEX root = mobicom2018.tex

\section{Discussion}
\label{sec.diss}

\vspace{0mm}	
\noindent	
\textbf{Impact on Mobile Vision Systems.} 
{\sysname} represents the first framework that supports resource-aware multi-tenant on-device deep learning for mobile vision systems.
%Automatically massive deployment.
%
This is achieved by replacing fixed resource-accuracy trade-offs with flexible ones and dynamically selecting the optimal trade-off at runtime to deliver the maximum performance subject to resource constraint.
%In practice, when releasing a deep learning-based mobile application, a trained model may need to be deployed on numerous heterogeneous hardware platforms with different resources.
%However, it is cumbersome to train specific model for each device.
%Since {\sysname} addresses some unique challenges in continuous mobile vision, 
We envision {\sysname} could become a useful framework for future mobile vision systems.

%eliminates this problem by providing a self-contained model. When deploying a new model to a new device, it can automatically switch to the suitable descendant model to deliver the maximum performance subject to the resource constraints.

%%
\vspace{0.5mm}	
\noindent	
\textbf{Generality of {\sysname}.}
In this work, we selected VGG Net and ResNet as two representative deep learning models to implement and evaluate {\sysname}.
However, {\sysname} can be generalized to support many other popular deep learning models such as GoogLeNet \cite{szegedy2015going}, MobileNets \cite{howard2017mobilenets}, and BinaryNet \cite{courbariaux2016binarized}.
%We expect {\sysname} would also benefit many other popular DNN models such as GoogLeNet (i.e., Inception Network), SqueezeNet, and BinaryNet.
%This is because these DNN models are built on top of VGG and ResNet and thus have very similar architectures. 
%Thus, we expect that NestDNN can be beneficial to a wide range of DNN models. 
%
{\sysname} can also be generalized to support other computing tasks beyond computer vision tasks. 
In this sense, {\sysname} is a generic framework for achieving resource-aware multi-tenant on-device deep learning.

\vspace{0.5mm}	
\noindent	
\textbf{Limitation.}
{\sysname} in the current form has some limitation.
%
%In Figure 3, we have compared the inference accuracies achieved by the DNN model when it is pruned by TRR and L1-norm. As shown, the accuracy achieved by TRR is higher than L1-norm. Based on this observation, we expected that the end-to-end inference performance achieved by TRR should be higher than L1-norm, and thus we only reported the end-to-end performance achieved by TRR. 
%
Although our TRR approach outperforms the $\mathcal{L}1$-norm approach in filter pruning, its computational cost is much higher than $\mathcal{L}1$-norm. 
This significantly increases the cost of the multi-capacity model generation process, even though the process is only performed once. 
We will find ways to reduce the cost and leave it as our future work.

%Even though {\sysname} is focused on continuous mobile vision, many proposed methods can be generalized.
%First of all, the proposed filter pruning method can be applied to DNN models, enabling efficient on-device mobile deep learning and further push the development of IoT.
%Second, the proposed freeze-grow training paradigm can be used to train a highly self-contained DNN model to get rid of the dependence on cloud-based services.
%Third, in this work, we investigate {\sysname} on a single end device. However, {\sysname} can be also used at the edge or even in the cloud. Edge devices and cloud servers are powered by GPU and can afford relatively intensive computation. Equipped with {\sysname}, they can provide more flexible inference queries for end devices.

%%
%\vspace{1mm}	
%\noindent	
%\textbf{Opportunities for Improvement.}
%We proposed {\sysname} and implement a proof-of-concept system. 
%There are opportunities for improving {\sysname}.
%On one hand, other complementary compression techniques can be combined to generated a more compact and efficient model for inference.
%On the other hand, we can implement caching techniques to further reduce switching overhead, such as \textit{Least Recently Used}.
%We leave them as our future work.

%
%It is possible that descendant models generated by {\sysname} still contain redundant weights.
%
%To solve this issue, methods of pruning redundant weights must be developed.
%
%As an tentative solution, after smallest weights have been pruned, all the descendant models can be fine-tuned jointly. We leave it as our future work.

\vspace{-1.5mm}
%!TEX root = mobicom2018.tex

\section{Related Work}
\label{sec.rw}

%%%%%%% TPCs
% Lin Zhong; Nic Lane; Mattai; Youngki Lee.

%%
%\noindent
%{\sysname} is related to two research areas: 1) deep learning model compression; and 2) continuous mobile vision.
%In this section, we brief review the most relevant work within each area and compare them with our work.

\vspace{0mm}
\noindent
\textbf{Deep Neural Network Model Compression.}
% 1. summarize all the existing work in this area.
% 2. explain what are the new contributions of our work compared to the existing work.
%
%
Model compression for deep neural networks has attracted a lot of attentions in recent years due to the imperative demand on running deep learning models on mobile systems.
% due to their limited available resources compared to cloud platforms. 
One of the most prevalent methods for compressing deep neural networks is pruning. 
%
%Due to the imperative demand for running DNN models on resource-constrained mobile systems, many DNN model compression approaches have been proposed.
%
Most widely used pruning approaches focus on pruning model parameters~\cite{han2015deep, han2015learning}. 
Although pruning parameters is effective at reducing model sizes, it does not necessarily reduce computational costs~\cite{luo2017thinet, molchanov2016pruning, li2016pruning}, making it less useful for mobile systems which need to provide real-time services. 
%In contrast, our approach focuses on pruning filters, which effectively reduces the size of a deep learning model and its computational cost.
%Han \textit{et al.}~\cite{han2015learning} proposed a parameter pruning method that removes node connections with small weights. 
%Although this method is effective at reducing model sizes, it does not effectively reduce computational costs.
%, making it less useful for mobile systems which need to provide real-time services.
%
To overcome this problem, Li \textit{et al.}~\cite{li2016pruning} proposed a filter pruning method that has achieved up to $38\%$ reduction in computational cost.
%In contrast, 
%
Our work also focuses on compressing deep neural networks via filter pruning. 
Our proposed filter pruning approach outperforms the state-of-the-art.
%However, we have proposed a novel filter importance ranking approach that outperforms state-of-the-art filter pruning method.
%
Moreover, unlike existing model compression methods which produce pruned models with fixed resource-accuracy trade-offs, our proposed multi-capacity model is able to provide dynamic resource-accuracy trade-offs.
%{\sysname} conducts model compression dynamically based on the available runtime resources. 
%
This is similar to the concept of dynamic neural networks in the deep learning literature~\cite{guo2016dynamic,lin2017runtime,huang2017densely}.

\vspace{0.5mm}
\noindent
\textbf{Continuous Mobile Vision.}
% 1. summarize all the existing work in this area.
% 2. explain what are the new contributions of our work compared to the existing work.
%In recent years, there is an increasing demand of continuous vision analysis in either larger clusters or mobile devices. Therefore, researchers have been focusing on system optimization when multiple continuous vision analysis happen simultaneously. 
%
%%
The concept of continuous mobile vision was first advanced by Bahl \textit{et al.}~\cite{bahl2012vision}.
The last few years have witnessed many efforts towards realizing the vision of continuous mobile vision~\cite{likamwa2015starfish, mcdnn, mathur2017deepeye, deepmon}.
In particular, in~\cite{likamwa2015starfish}, LiKamWa \textit{et al.} proposed a framework named \textsf{Starfish}, which enables efficient running concurrent vision applications on mobile devices by sharing common computation and memory objects across applications.
Our work is inspired by \textsf{Starfish} in terms of sharing.
By sharing parameters among descendent models, our proposed multi-capacity model has a compact memory footprint and incurs little model switching overhead.
Our work is also inspired by~\cite{mcdnn}.
In~\cite{mcdnn}, Han \textit{et al.} proposed a framework named \textsf{MCDNN}, which applies various model compression techniques to generate a catalog of model variants to provide different resource-accuracy trade-offs.
%It also proposed a scheduling method for solving multiple deep learning classification problems requests under resource constraints.
%
However, in \textsf{MCDNN}, the generated model variants are independent of each other, and it relies on cloud connectivity to retrieve the desired model variant.
In contrast, our work focuses on developing an on-device deep learning framework which does not rely on cloud connectivity.
Moreover, \textsf{MCDNN} focuses on model sharing across concurrently running applications. 
In contrast, {\sysname} treats each of the concurrently running applications independently, and focuses on model sharing across different model variants within each application.

\vspace{-1.5mm}
%!TEX root = mobicom2018.tex

\section{Conclusion}
\label{sec.con}
%\noindent
%%
In this paper, we presented the design, implementation and evaluation of {\sysname}, a framework that enables resource-aware multi-tenant on-device deep learning for mobile vision systems. 
{\sysname} takes the dynamics of runtime resources in a mobile vision system into consideration, and dynamically selects the optimal resource-accuracy trade-off and resource allocation for each of the concurrently running deep learning models to jointly maximize their performance.
%It contributes novel techniques that address the unique challenges of mobile vision systems.
%
%\xiao{can we add some comment to show the calculation of these numbers? for faster verification}
%
%Based on our freeze-and-grow training paradigm, {\sysname} achieves up to $7.2\times$ computational cost reduction and $8.6\times $ model size reduction without accuracy drop.
%Based on our resource-aware scheduler, {\sysname} achieves up to $2.1\times $ FPS and $2.2\times $ energy consumption reduction on concurrent mobile applications.
%
%We have also shown that {\sysname} achieves as much as $3.1\times $ energy consumption reduction caused by model switching.
%
We evaluated {\sysname} using six mobile vision applications that target some of the most important vision tasks for mobile vision systems.
%These applications are developed based on two widely used deep learning models -- VGGNet and ResNet -- and six commonly used datasets from computer vision community.
%
Our results show that {\sysname} outperforms the resource-agnostic status quo approach in inference accuracy, video frame processing rate, and energy consumption.
We believe {\sysname} represents a significant step towards turning the envisioned continuous mobile vision into reality.

\vspace{-5mm}
%!TEX root = mobicom2018.tex

\section{Acknowledgement}
\label{sec.ack}
\noindent
We thank the anonymous shepherd and reviewers for their valuable feedback. 
This work was partially supported by NSF Awards CNS-1617627, IIS-1565604, and PFI:BIC-1632051.

%We are grateful to NVIDIA Corporation for the donation of GPUs to support our research.

\bibliographystyle{ACM-Reference-Format}
\bibliography{sigproc}

\end{document}